%% file: main.tex
\documentclass[runningheads]{llncs}
\usepackage[T1]{fontenc}
\usepackage{graphicx}
\usepackage{color}

\usepackage{subcaption}
\usepackage[acronym]{glossaries}
\input{acronyms.tex}
\begin{document}
\title{Data-Centric Machine Learning for Earth Observation: Necessary and Sufficient Features}
\titlerunning{Data-Centric ML for EO}

\author{Hiba Najjar\inst{1,2}\orcidID{0000-0002-7498-794X} \and
Marlon Nuske\inst{2}\orcidID{0000-0002-0651-0664} \and
Andreas Dengel\inst{1,2}\orcidID{0000-0002-6100-8255}}
\authorrunning{H. Najjar et al.}

\institute{University of Kaiserslautern-Landau (RPTU), Kaiserslautern, Germany \\
\email{najjar@rptu.de} \and
German Research Center for Artificial Intelligence (DFKI), Kaiserslautern, Germany \\
\email{\{marlon.nuske,andreas.dengel\}@dfki.de}}
\maketitle            

\input{sections/00_asbtract}
\input{sections/01_intro}

\input{sections/02_method}

\input{sections/03_results}

\input{sections/04_conclusion}

\input{sections/99_credits}

\bibliographystyle{splncs04}
\bibliography{refs/related_work, refs/models, refs/xai_methods, refs/datasets}
\end{document}

%% file: acronyms.tex
\newacronym{ml}{ML}{Machine Learning}
\newacronym{ai}{AI}{Artificial Intelligence}
\newacronym{dl}{DL}{Deep Learning}
\newacronym{xai}{XAI}{eXplainable AI}

\newacronym{eo}{EO}{Earth Observation}
\newacronym{rs}{RS}{Remote Sensing}
\newacronym{rgb}{RGB}{Red-Green-Blue}
\newacronym{dem}{DEM}{digital elevation model}
\newacronym{srtm}{SRTM}{Shuttle Radar Topography Mission}
\newacronym{aster}{ASTER}{Advanced Spaceborne Thermal Emission and Reflection Radiometer}
\newacronym{alos}{ALOS}{Advanced Land Observing Satellite}
\newacronym{ecmwf}{ECMWF}{European Center for Medium-Range Weather Forecasts}
\newacronym{sar}{SAR}{synthetic aperture radar}
\newacronym{uav}{UAV}{unmanned aerial vehicle}
\newacronym{lidar}{LiDAR}{Light Detection and Ranging}
\newacronym{fapar}{FAPAR}{Fraction of Absorbed Photosynthetically Active Radiation}

\newacronym{ndvi}{NDVI}{normalized difference vegetation index}
\newacronym{evi}{EVI}{enhanced vegetation index}

\newacronym{rf}{RF}{random forest}
\newacronym{svm}{SVM}{support vector machine}
\newacronym{mlp}{MLP}{multilayer perceptron}
\newacronym{dnn}{DNN}{deep neural network}
\newacronym{cnn}{CNN}{convolutional neural network}
\newacronym{fnn}{FNN}{feed forward neural network}
\newacronym{rnn}{RNN}{recurrent neural network}
\newacronym{lstm}{LSTM}{Long short-term memory}
\newacronym{gru}{GRU}{Gated Recurrent Unit}
\newacronym{tae}{TAE}{Temporal Attention Encoder}
\newacronym{ltae}{L-TAE}{Lightweight-TAE}
\newacronym{tempcnn}{TempCNN}{Temporal Convolutional Neural Networks}

\newacronym{cam}{CAM}{Class Activation Mapping}
\newacronym{gradcam}{Grad-CAM}{Gradient-weighted Class Activation Mapping}
\newacronym{shap}{SHAP}{SHapley Additive exPlanations}
\newacronym{lime}{LIME}{Local Interpretable Model-agnostic Explanation}
\newacronym{svs}{SVS}{Shapley Value Sampling}
\newacronym{sgs}{SG-SQ}{SmoothGrad-Squared}
\newacronym{var}{VAR}{VarGrad}
\newacronym{gb}{GB}{Guided Backprop}
\newacronym{roar}{ROAR}{RemOve And Retrain}

\newacronym{r2}{$R^2$}{coefficient of determination}

%% file: sections/00_asbtract.tex
\begin{abstract}


The availability of temporal geospatial data in multiple modalities has been extensively leveraged to enhance the performance of machine learning models. While efforts on the design of adequate model architectures are approaching a level of saturation, focusing on a data-centric perspective can complement these efforts to achieve further enhancements in data usage efficiency and model generalization capacities.
This work contributes to this direction.  We leverage model explanation methods to identify the features crucial for the model to reach optimal performance and the smallest set of features sufficient to achieve this performance. We evaluate our approach on three temporal multimodal geospatial datasets and compare multiple model explanation techniques. Our results reveal that some datasets can reach their optimal accuracy with less than 20\% of the temporal instances, while in other datasets, the time series of a single band from a single modality is sufficient.

\keywords{Data-Centric Machine Learning \and Explainable AI \and Multi-modal data \and Time Series}
\end{abstract}

%% file: sections/01_intro.tex
\section{Introduction}\label{sec:intro}

        The abundance of \gls{eo} data presents a significant opportunity for leveraging \gls{ml} to train highly accurate models for \gls{eo}-related applications in real-world scenarios. The data used is usually derived from various sources such as satellites, aerial imagery, and ground-based sensors, including time series and static features. This diversity provides rich datasets that can be harnessed by multi-modal learning techniques to improve the predictive capacities of \gls{ml} models \cite{mena2024common}. 

        Most recent work in the \gls{eo} research community has focused on enhancing model architectures and training strategies to boost performance, and significant advancements have been made. However, as a complementary component to these model-centric efforts, research is increasingly diverging towards a more data-centric approach. This shift aims to better address challenges faced during data acquisition and data curation stages. Subsequently, the enhancement of the quality of the input data can improve the performance and reliability of the models. Furthermore, incorporating a feedback loop that includes model evaluation results can provide valuable insights for refining both the data and the models.

        Within the scope of data-centric machine learning, this work focuses on feature engineering techniques. In particular, feature selection methods aim at identifying the most useful and necessary features relevant to the task at hand. The objective is to avoid supplying the model with redundant information or extraneous features in the available data that may cause the model to learn spurious correlations and hinder its capacity to generalize. The prevailing belief that ”more data is better” does not always hold true, and often comes at the expense of immense computational resources usage and a heavy impact on the environment \cite{falk2023challenging}.

        To enhance the feature selection process, we employ \gls{xai} techniques to guide the selection approach. In particular, we leverage feature attribution methods to estimate how much each feature contributes to the final predictions. By employing an incremental deletion approach, we iteratively remove less important features until an optimal set of predictive features is identified. This method ensures an active optimization of the data at each cycle, leading to a final model that efficiently utilizes the available data.

    Traditional feature selection methods, such as filter, wrapper, and embedded approaches, often involve exhaustive search strategies that can be computationally prohibitive for large datasets. 
        \textbf{Filter methods} are preprocessing steps independent of the model training. They select features based on their statistical properties and relevance to the target variable.
        \textbf{Wrapper methods} evaluate feature subsets based on the performance of a specific machine learning algorithm. 
            One prominent work in this domain is the Recursive Feature Elimination (RFE) algorithm, which recursively removes the least significant features based on their importance weights, as determined by a support vector machine (SVM) \cite{rfe_guyon2002gene}. 
            In \gls{eo}, Zhang et al. \cite{zhang2023marine} use a similar technique based on feature importance extracted from a random forest model to identify the most relevant features in marine data. A major limitation of this approach is that the feature importance is extracted from a model different from the main one, while these scores are usually model dependent.
        \textbf{Embedded methods} perform feature selection during the model training process, often through regularization techniques. Examples include the Least Absolute Shrinkage and Selection Operator (LASSO) method \cite{lasso_tibshirani1996regression}, which performs feature selection by enforcing sparsity through L1 regularization, and implicitly removes less important features during model training.
    
    The incremental deletion approach is also used in the field of \gls{xai} to evaluate the correctness of feature importance scores, i.e. how faithful these scores are to the model \cite{nauta2023anecdotal}. Given that deleting features by setting them to zero, for instance, can lead to out-of-distribution samples, an alternative is to retrain the model on the modified data. Specifically, Hooker et al. \cite{hooker2019benchmark} propose the \gls{roar} method, which evaluates the feature attribution estimates by removing the k\% most important features, and then retraining the model.
    
    In this paper, we use the \gls{roar} method to identify the set of sufficient and necessary predictive features in temporal multi-source geospatial data. Unlike the original study, which replaced the deleted features with a constant value, we completely remove the feature before retraining the model. This setup allows the simulation of a scenario where the deleted features are genuinely absent from the initial dataset. First, we select a highly performant model by comparing recurrent, convolutional, and attention-based networks. Subsequently, we incrementally delete either the most or least important features based on importance scores estimated by six different methods and retrain the selected model. We conduct separate feature deletion operations on the spectral and temporal dimensions to identify the necessary modalities and time steps, respectively, and validate our approach on multiple temporal and multi-modal geospatial datasets.

%% file: sections/02_method.tex
\section{Methodology}\label{sec:method}

    \subsection{Datasets}\label{subsec:data}
        We apply the incremental deletion framework to the following three \gls{eo} datasets, which are composed of different modalities and span multiple years, including both regression and classification tasks.
        
        \paragraph{CropHarvest.} The CropHarvest dataset is a multi-source temporal dataset designed for crop classification tasks \cite{tseng2021cropharvest}. The input modalities include satellite data from Sentinel-2 and Sentinel-1, weather time series, and static topographic information. The temporal modalities are provided on a monthly basis over multiple years (2016 - 2022). We utilize a multi-crop version of this dataset, which includes data for 10 classes.

        \paragraph{CropYield.} For crop yield estimation, we use a dataset specifically aimed at predicting cereal crop yield \cite{perich2023pixel}. This regression task involves predicting the amount of crop (in tonnes per hectare, t/ha) grown in a particular location during the growing seasons from 2017 until 2021. The input modalities for this dataset include multispectral satellite data from Sentinel-2 and weather time series. The weather data is aligned with the satellite temporal resolution of five days, covering the period from seeding to harvesting.

        \paragraph{China PM2.5.}  The China PM2.5 dataset provides data for the prediction of PM2.5 concentration levels across various regions in China to track air pollution \cite{pm25}. This dataset covers the years from 2010 until 2015 and includes multiple modalities, namely weather information and ground-based air quality measurements. The task is to predict particulate matter concentrations using hourly measurements from the preceding 7 days leading up to the date of the target value. 
        
\subsection{Modeling}\label{subsec:models}
        Each dataset is split into three subsets: the training set comprises data from all years except the last two, which are uniformly divided into validation and test sets. This strategic split aims at reproducing the case where real-life application can only train the models on data from previous years to be deployed for upcoming seasons.
       
        To identify the most effective architecture for processing multivariate time series data, we conduct a comparative evaluation of seven different model architectures. Each architecture is tested under multiple hyperparameter settings and evaluated on the validation set to determine its performance. The models compared in this study include a \gls{mlp}, a basic \gls{rnn} and two variants: \gls{lstm} \cite{lstm_hochreiter1997long} and \gls{gru} \cite{gru_cho2014properties}, a \gls{tempcnn}, and finally two attention-based models: \gls{tae} \cite{tae_garnot2020satellite} and \gls{ltae} \cite{ltae_garnot2020lightweight}.

    \subsection{Attribution estimators}

        Feature attribution methods are \gls{xai} tools that aim to quantify the contribution of input features to the output of a machine learning model. These methods provide insights into which features are influential in making predictions. 
        In this study, we compare perturbation-based and gradient-based methods to infer feature attributions, namely \gls{svs} \cite{svs_strumbelj2010efficient} and \gls{gb} \cite{gb_springenberg2015striving} methods.
        Additionally, two ensemble-based variants of each base estimator are implemented: \gls{sgs} \cite{sg_smilkov2017smoothgrad} and \gls{var} \cite{var_adebayo2018sanity}. As shown in \cite{hooker2019benchmark}, these ensembling methods can strongly improve the correctness of the attributions of the base estimators.

        To ensure that the analysis captures the overall influence of each feature, the attributions are considered in their absolute values.
        A random selection of 5000 samples from the training set is used to estimate the attributions, which are averaged to derive the ranking of the features. The same selection of samples is used across all experiments of the same dataset.

        We employ a feature grouping strategy applicable only on the \gls{svs} method to estimate temporal and band importance, by considering collective contributions to the model's predictions. Temporal importance is estimated by grouping the bands at each time step, perturbing them together to infer the significance of that particular time step. Similarly, band importance is assessed by treating the time series of each band as a group.

    \subsection{Incremental Deletion}

        \paragraph{Baseline Model}
        - After evaluating multiple architectures as described in \ref{subsec:models}, the model with the best metric score on the validation set is selected as the baseline model for the incremental deletion cycles. The metric score is defined based on the optimization task, such as the accuracy for classification or the \gls{r2} for regression. The validation loss is used for early stopping. 
        
        \paragraph{Deletion Order}
        - We conduct two types of feature deletions: either based on the most important features or based on the least important features. 
        Progressively deleting the most important features can reveal the necessity of the key features to reach the baseline performance. 
        Conversely, eliminating features that do not significantly contribute to the predictions can reduce noise in the input data, potentially enhancing model performance. This method can also identify a subset of features that are sufficient to reach the baseline performance after all extraneous features have been eliminated.
        The number of features deleted in each step depends on the dataset. By default, the deletion process addresses a single feature at a time. For long time-series, a larger step is used.

        \paragraph{Cycles}
        - After training the baseline model, feature attributions are estimated using six different estimators, grouped by time-steps or bands, and the corresponding features are ranked accordingly. A new copy of the dataset is created by deleting the most or least important features, and a new model instance is trained with this modified data. The model architecture remains consistent, except for the modifications necessary to handle the new input size.
        For instance, removing a spectral band from the satellite modality would require adjusting the number of input channels in the first convolutional layer.
        Post-training, the new model is explained, and the attribution scores averaged over the selected samples are used to rank the features and decide which ones to delete in the next cycle. This process is repeated, updating the feature attribution estimates and modifying the input data after each training, until only a single (or a set of) feature(s) is left in the addressed dimension at the final cycle.

%% file: sections/03_results.tex
\section{Results}\label{sec:res}

    \subsection{Model selection}\label{subsec:res_model}
        
            \input{tables_figures/modeling_results}
        
        We trained each model architecture with various configurations and compared their performance on the validation set. The metrics reported in Table \ref{tab:model_res} include accuracy for the classification task and the \gls{r2} for regression tasks, for the best-performing configuration of each model architecture.
        
        The results indicate that \gls{tempcnn} consistently achieves the highest scores across all datasets, demonstrating its efficiency in handling both short and long time-series data.
        Attention-based models achieve the second-best scores, followed closely by the \gls{gru} network. 
        Interestingly, the \gls{mlp} outperformed the \gls{lstm} network in all datasets, as well as the \gls{rnn} in the two regression tasks.

    \subsection{Incremental deletion}\label{subsec:cyclicxai}

        We apply the incremental deletion approach on each dataset using the \gls{tempcnn} architecture. We evaluate the model after each cycle and report its performance results for band and time-step deletion on the validation set, as shown in Figures \ref{fig:band_del} and \ref{fig:ts_del}, respectively. A horizontal line indicates the baseline performance in each plot.

\input{tables_figures/band_deletion_figures}

\input{tables_figures/ts_deletion_figures}

        In the top row of Figure \ref{fig:band_del}, the most important bands are deleted first. We observe that the performance progressively declines in the CropHarvest and CropYield datasets. The decrease in performance is more significant for the PM2.5 dataset, especially when using the \gls{svs} attribution estimators to rank the features.
        This observation has two implications: first, the correctness of the attributions returned by \gls{svs} exceeds those provided by the \gls{gb} method; and second, the baseline performance in the PM2.5 dataset relies on the two most important features (wind speed and direction), with the remaining features being insufficient for the model to achieve the baseline performance. In contrast, in the agricultural datasets, many important features can be dropped before a significant decline in model performance is observed. This also indicates that these features are not necessary for achieving a performance comparable to the baseline.

        Deleting the least important bands, as shown in the second row of Figure \ref{fig:band_del}, reveals additional insights. In the CropHarvest dataset, removing up to 70\% of the least important bands does not reduce the model accuracy below 60\%.
        In the CropYield dataset, more than 80\% of the bands can be removed according to the \gls{svs} method, and the model can still recover its baseline performance. 
        Using the same methods, the model trained on the PM2.5 dataset maintains its baseline performance even when 65\% of the features are deleted. In this case, wind speed, wind direction, and humidity conditions were sufficient for achieving the baseline optimal performance.

        The results of the time-step deletion analysis provide insights into the time periods whose absence significantly impacts model performance and those which are sufficient to achieve baseline accuracy.
        In Figure \ref{fig:ts_del}, removing the most important instances first, as shown in the top row, results in a consistent decline in performance in the CropHarvest dataset, particularly when using the \gls{svs} estimator. For the PM2.5 dataset, the \gls{r2} scores with the \gls{svs} estimator are lower than those achieved with the \gls{gb} estimator, especially when more than 30\% of the time-steps are removed. The behavior of the two estimators is mixed in the CropYield dataset. In all datasets, the moderate slope of the curves indicates that the information required by the model for accurate prediction is distributed across multiple instances, rather than being concentrated in a few critical time-steps.

        The results of removing the least important instances first in the second row in Figure \ref{fig:ts_del} show that the model can still perform similarly to the baseline after deleting more than 30\% and 40\% of the time-steps in CropHarvest and CropYield, respectively. In PM2.5, a performance comparable to the baseline can be achieved even when more than 80\% of the instances are deleted, according to the feature ranks provided by \gls{svs}. Taking a closer look at the time-steps left at this point revealed that the hourly instances from the last two days  were sufficient to reach a high \gls{r2} score.

        We repeated these experiments using the ensemble-based variants. The results of the incremental deletion based on \gls{sgs} and \gls{var} methods applied to each estimator are illustrated in Figures \ref{fig:band_del_sg} and \ref{fig:ts_del_sg}. 

        In Hooker et al. \cite{hooker2019benchmark}, these variants significantly improved the correctness of gradient-based attribution estimator. In our experiments, the only noticeable improvement is observed in the PM2.5 dataset when removing the most important bands first. 
        Specifically, \gls{gb} (\gls{var}) shows a significant decline in model performance, reflecting a correct estimation of the feature importance ranking. 
        Another noteworthy improvement is observed in the CropYield dataset, where removing the least important time-steps first, as displayed in Figure \ref{fig:ts_del_sg}.(b), shows a high accuracy even in the last cycle of using the \gls{svs} (\gls{sgs}) method. This suggests a filtering of the least important features that is faithful to the model and its reasoning. Concretely, in this dataset, the first short wave infrared band from Sentinel-2 satellite data was sufficient for achieving baseline performance.

\input{tables_figures/band_deletion_figures_sg}

\input{tables_figures/ts_deletion_figures_sg}

%% file: tables_figures/modeling_results.tex
\begin{table}[t]
\caption{Best model performance for each architecture. We highlight in bold the best score in each dataset.}
\centering
\footnotesize
\begin{tabular}{cccccccccc}
\hline
 &  \textbf{ \quad } & \multicolumn{2}{c}{CropHarvest (Accuracy)} &  \textbf{ \quad } & \multicolumn{2}{c}{CropYield (\gls{r2})} &  \textbf{ \quad } & \multicolumn{2}{c}{China PM2.5 (\gls{r2})} \\
 &  & Validation & Test &  & Validation & Test &  & Validation & Test \\  \hline
MLP &  & 62.1 & 61.9 &  & 52.4 & 42.1 &  & 68.0 & 54.3 \\
RNN &  & 65.7 & 65.5 &  & 44.0 & 33.5 &  & 47.3 & 44.1 \\
LSTM &  & 56.8 & 56.9 &  & 43.3 & 35.6 &  & 63.0 & 57.1 \\
GRU &  & 63.6 & 63.6 &  & 54.6 & 47.8 &  & 71.5 & 64.1 \\
TempCNN &  & \textbf{87.8} & \textbf{86.6} &  & \textbf{67.4} & \textbf{61.8} &  & \textbf{77.5} & \textbf{67.2} \\
TAE &  & 67.9 & 67.9 &  & 58.5 & 51.0 &  & 66.3 & 57.5 \\
LTAE &  & 67.3 & 67.2 &  & 63.1 & 57.2 &  & 72.4 & 62.4 \\ \hline
\end{tabular}
\label{tab:model_res}
\end{table}

%% file: tables_figures/band_deletion_figures.tex
\begin{figure}[t]
    \centering
    \begin{tabular}{ccc}
        \begin{subfigure}{0.3\textwidth}
            \centering
            \includegraphics[width=\linewidth]{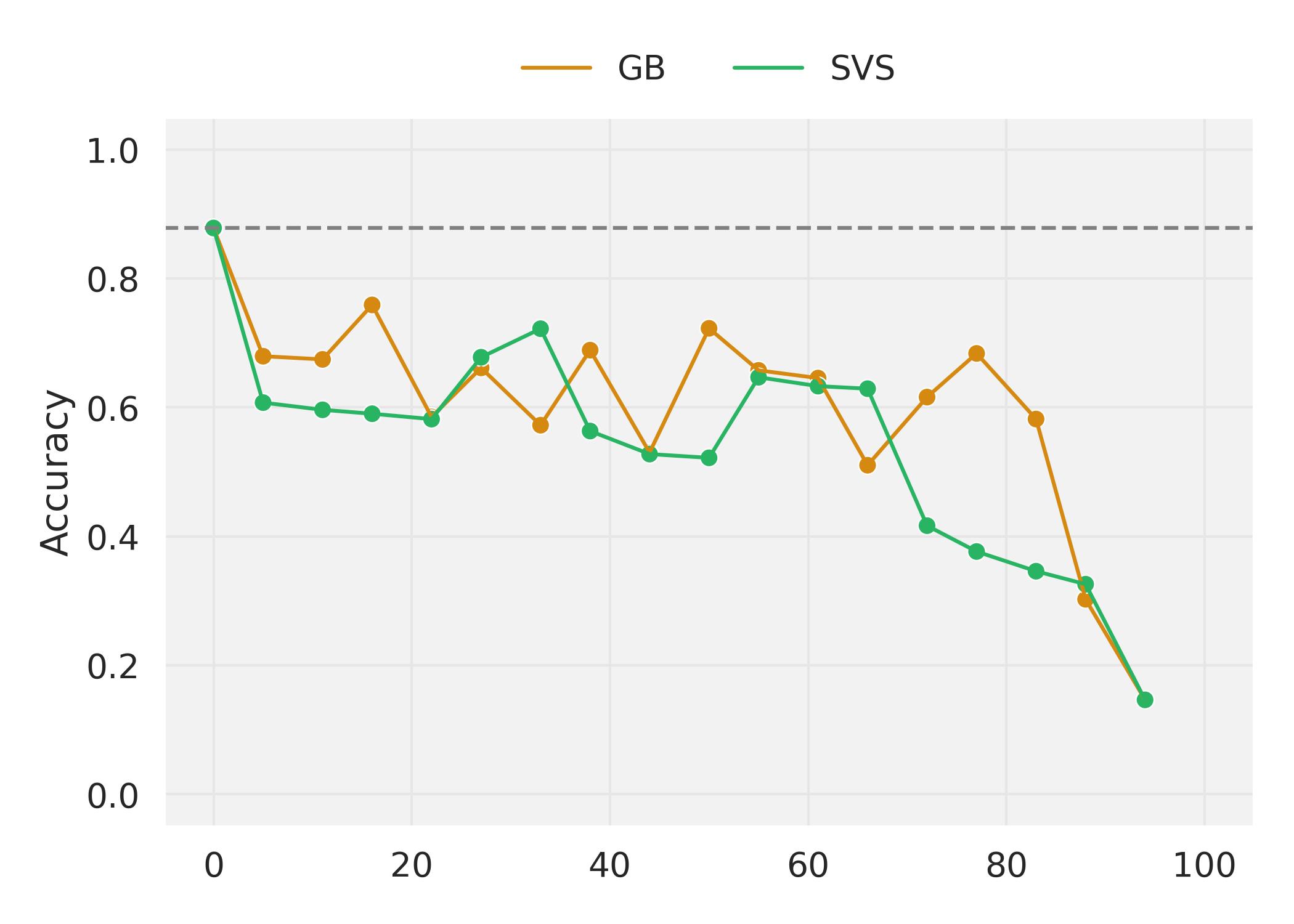}
        \end{subfigure} &
        \begin{subfigure}{0.3\textwidth}
            \centering
            \includegraphics[width=\linewidth]{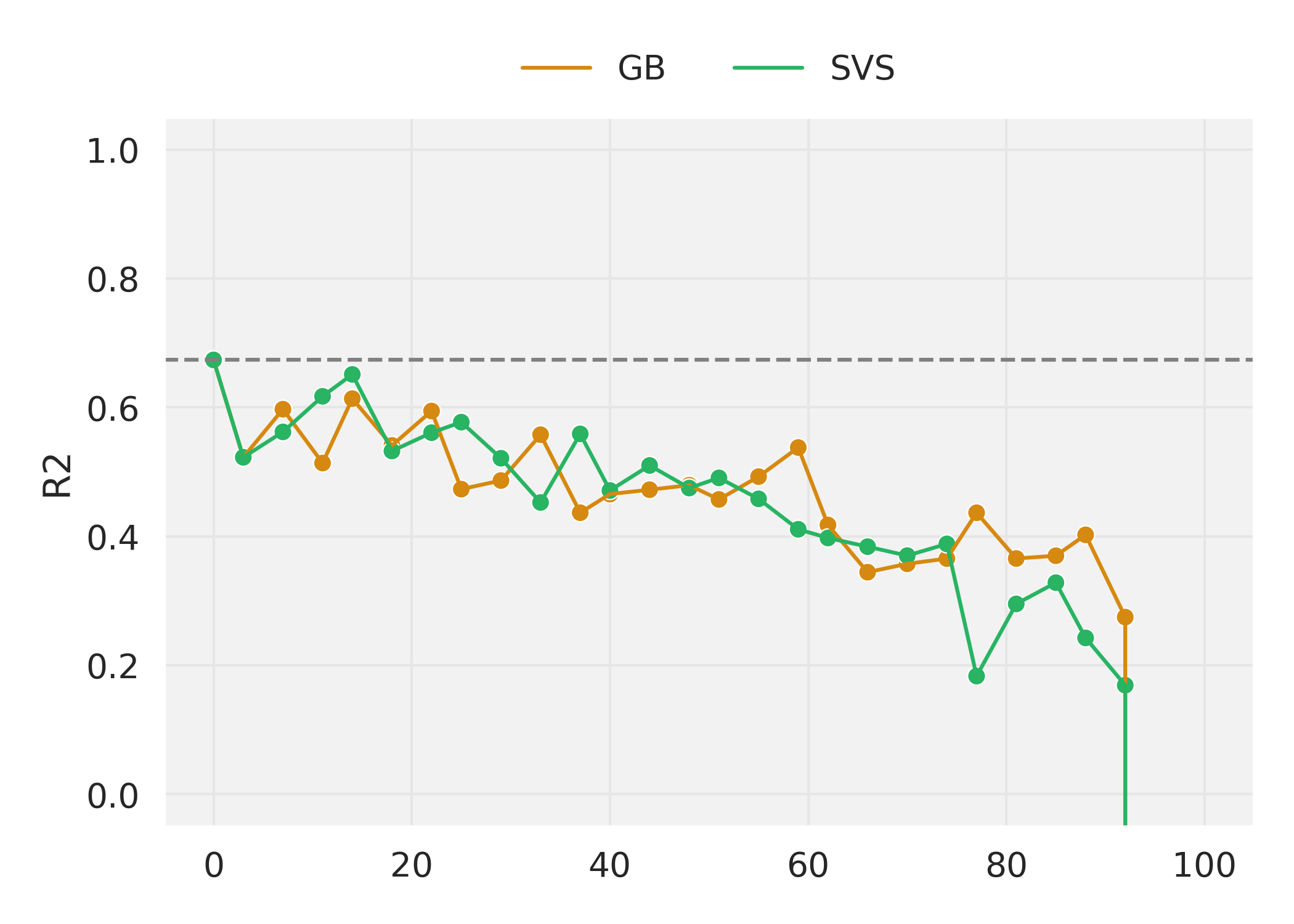}
        \end{subfigure} &
        \begin{subfigure}{0.3\textwidth}
            \centering
            \includegraphics[width=\linewidth]{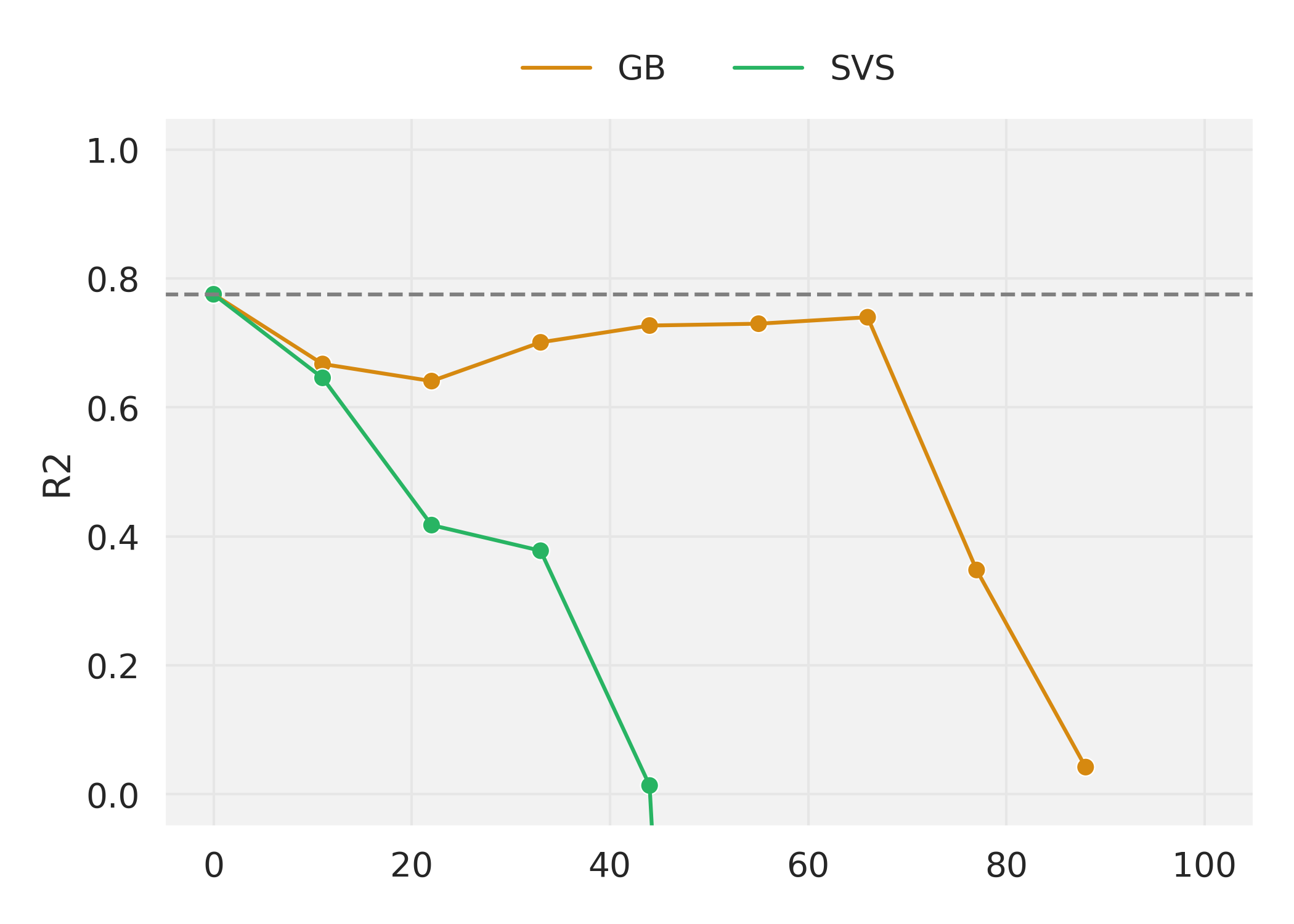}
        \end{subfigure} \\
        \begin{subfigure}{0.3\textwidth}
            \centering
            \includegraphics[width=\linewidth]{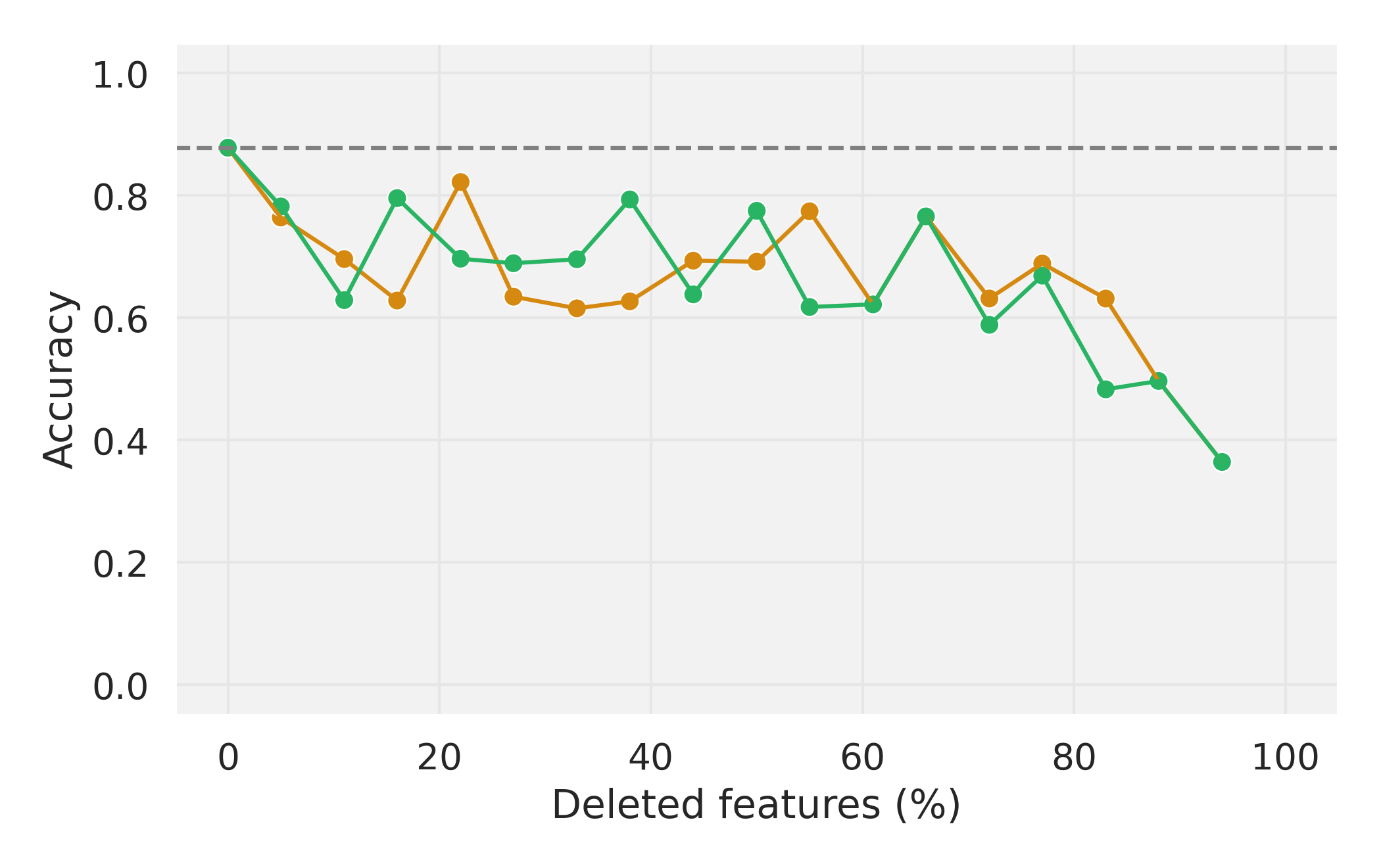}
            \caption{CropHarvest}
        \end{subfigure} &
        \begin{subfigure}{0.3\textwidth}
            \centering
            \includegraphics[width=\linewidth]{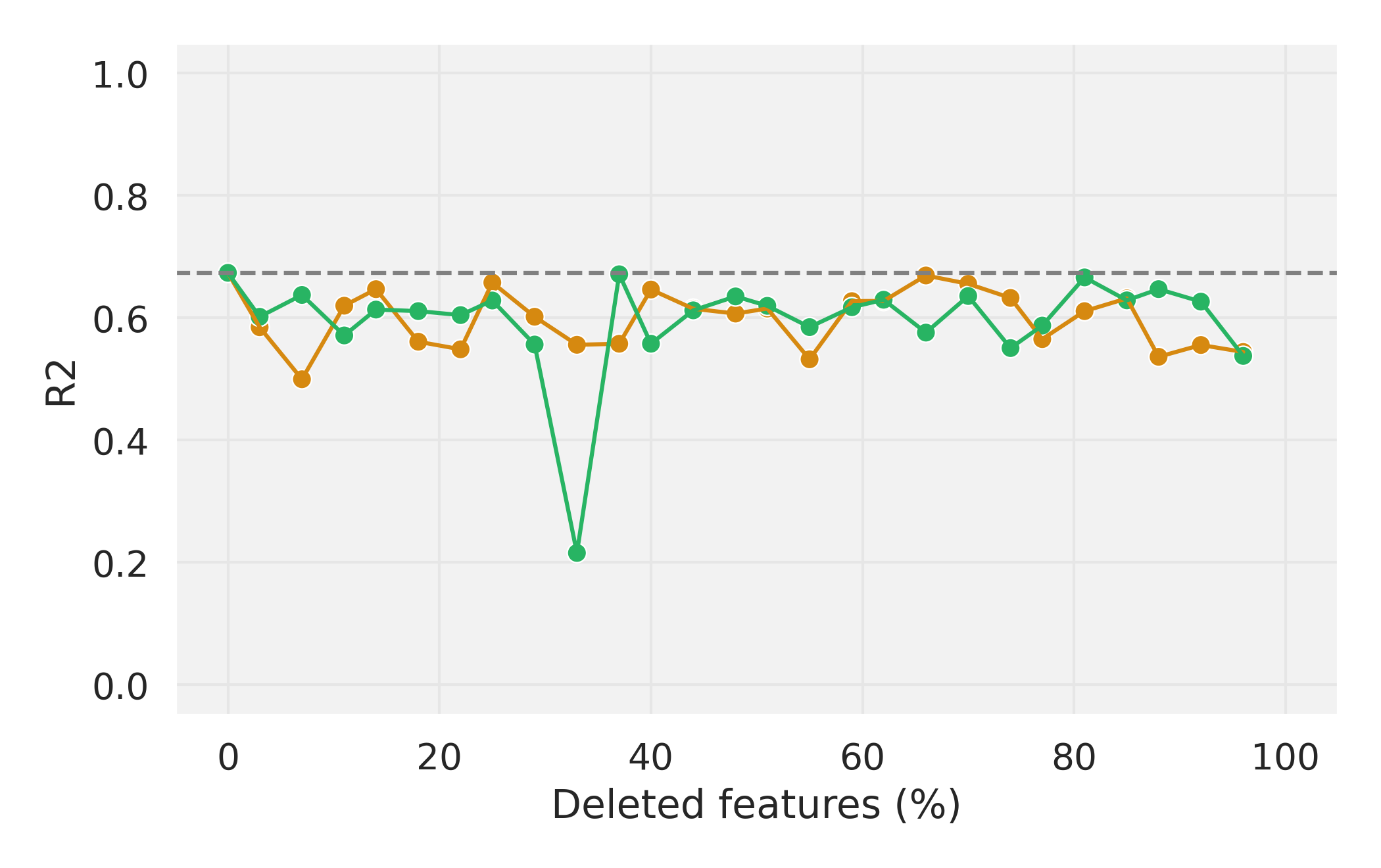}
            \caption{CropYield}
        \end{subfigure} &
        \begin{subfigure}{0.3\textwidth}
            \centering
            \includegraphics[width=\linewidth]{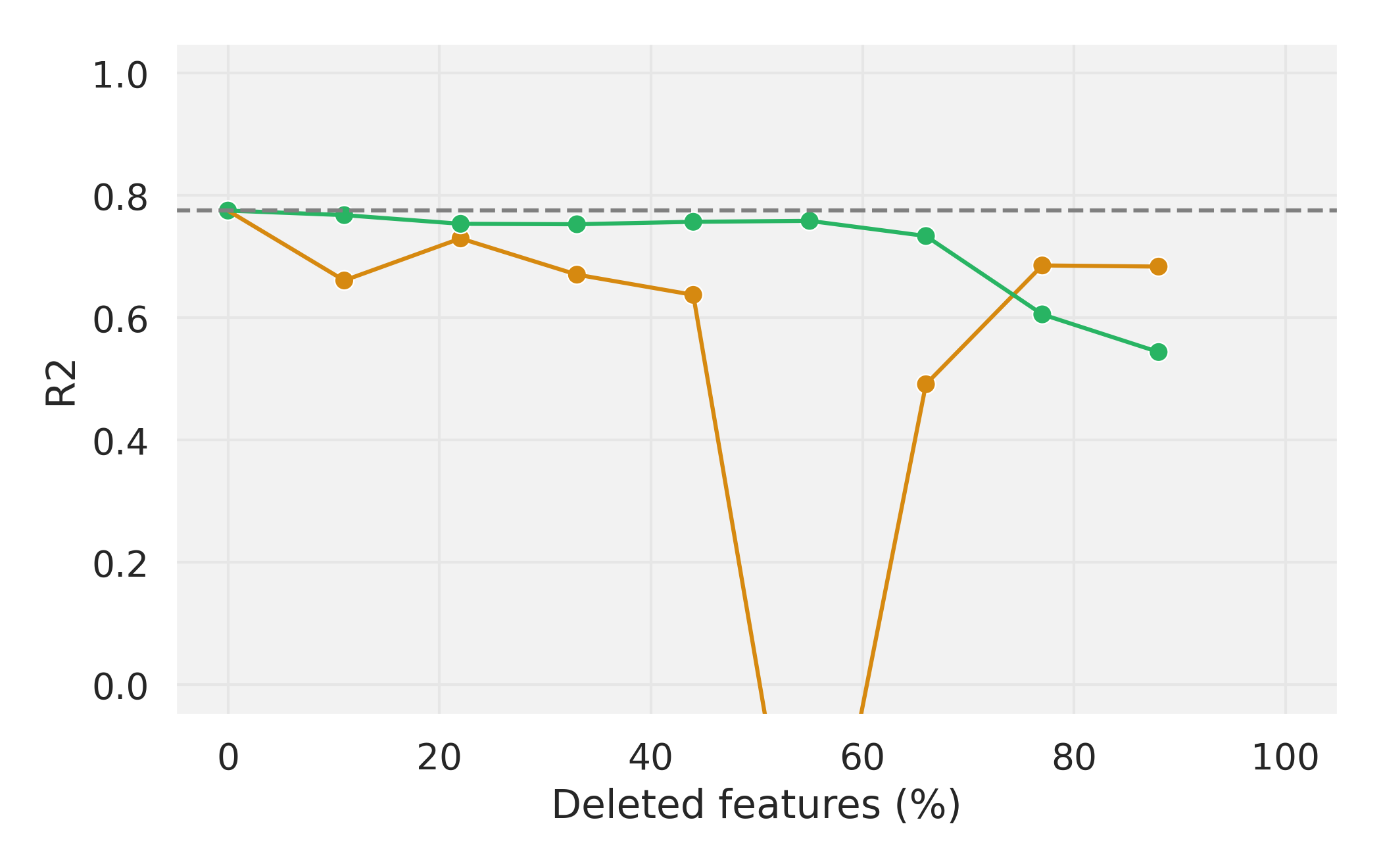}
            \caption{China PM2.5}
        \end{subfigure}
    \end{tabular}
    \caption{Incremental band deletion results, comparing  \gls{svs} and \gls{gb} attribution estimators. In the top row the most important bands are removed first, and the least important bands in the second row.}
    \label{fig:band_del}
\end{figure}

%% file: tables_figures/ts_deletion_figures.tex
\begin{figure}[ht]
    \centering
    \begin{tabular}{ccc}
        \begin{subfigure}{0.3\textwidth}
            \centering
            \includegraphics[width=\linewidth]{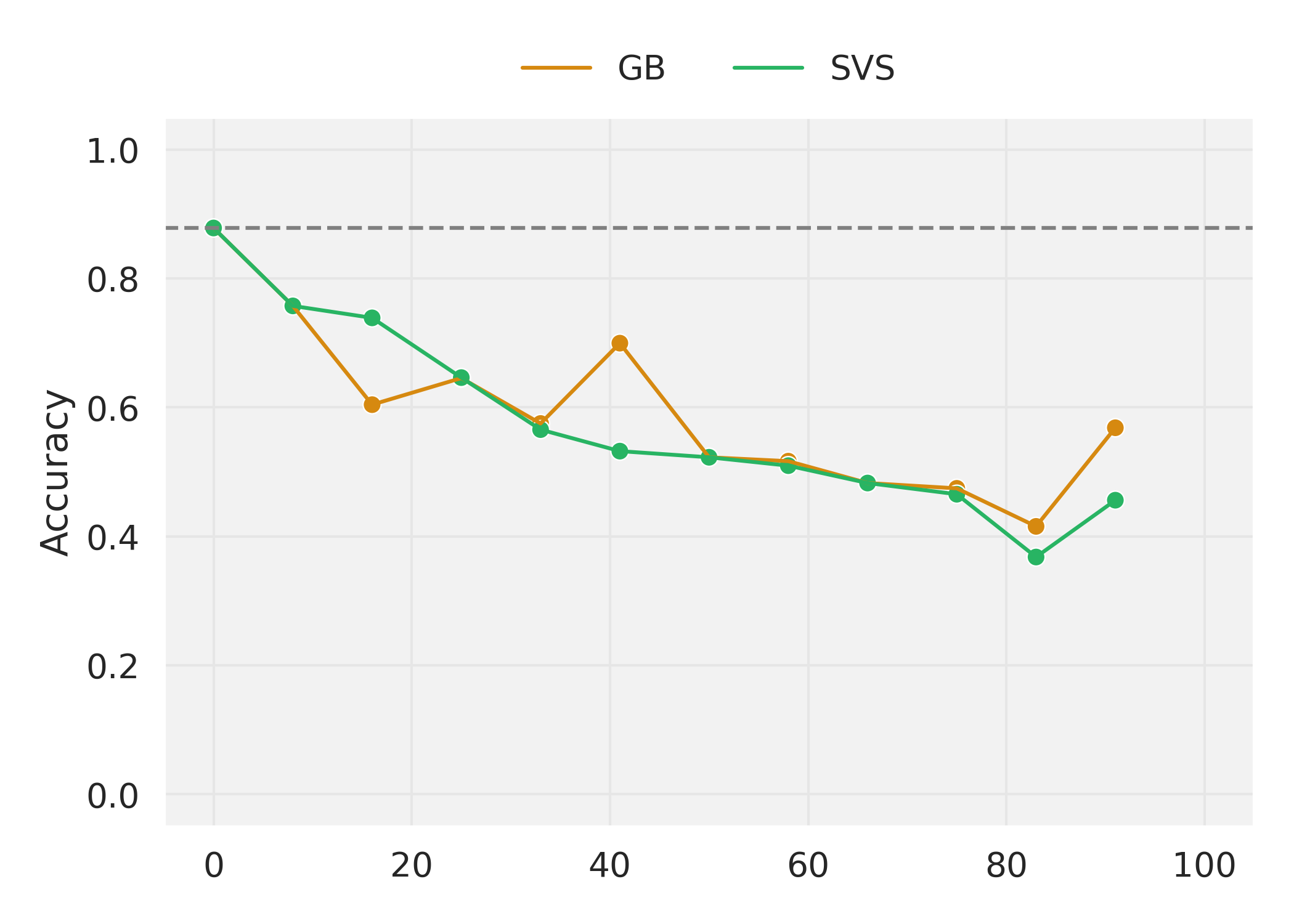}
        \end{subfigure} &
        \begin{subfigure}{0.3\textwidth}
            \centering
            \includegraphics[width=\linewidth]{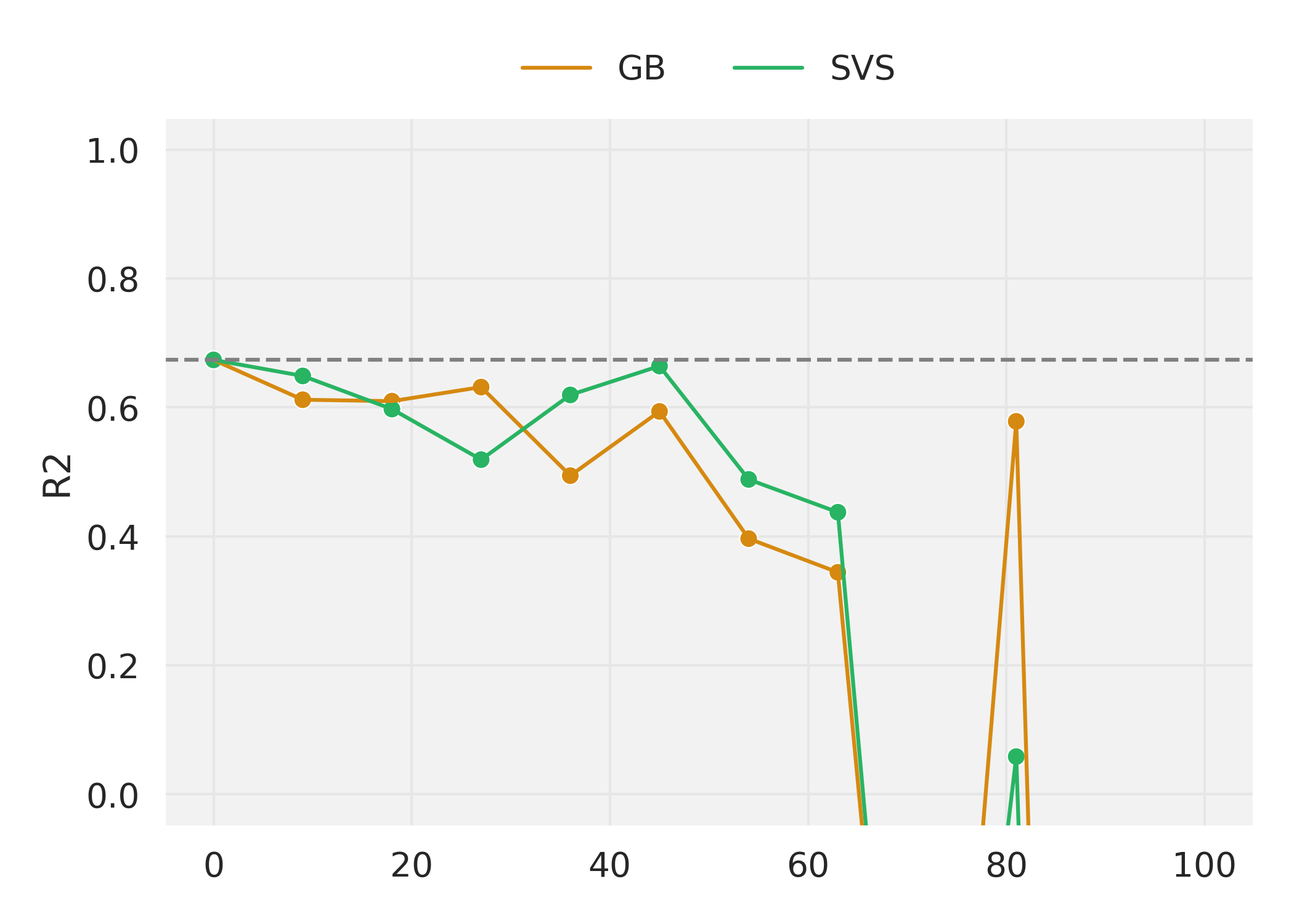}
        \end{subfigure} &
        \begin{subfigure}{0.3\textwidth}
            \centering
            \includegraphics[width=\linewidth]{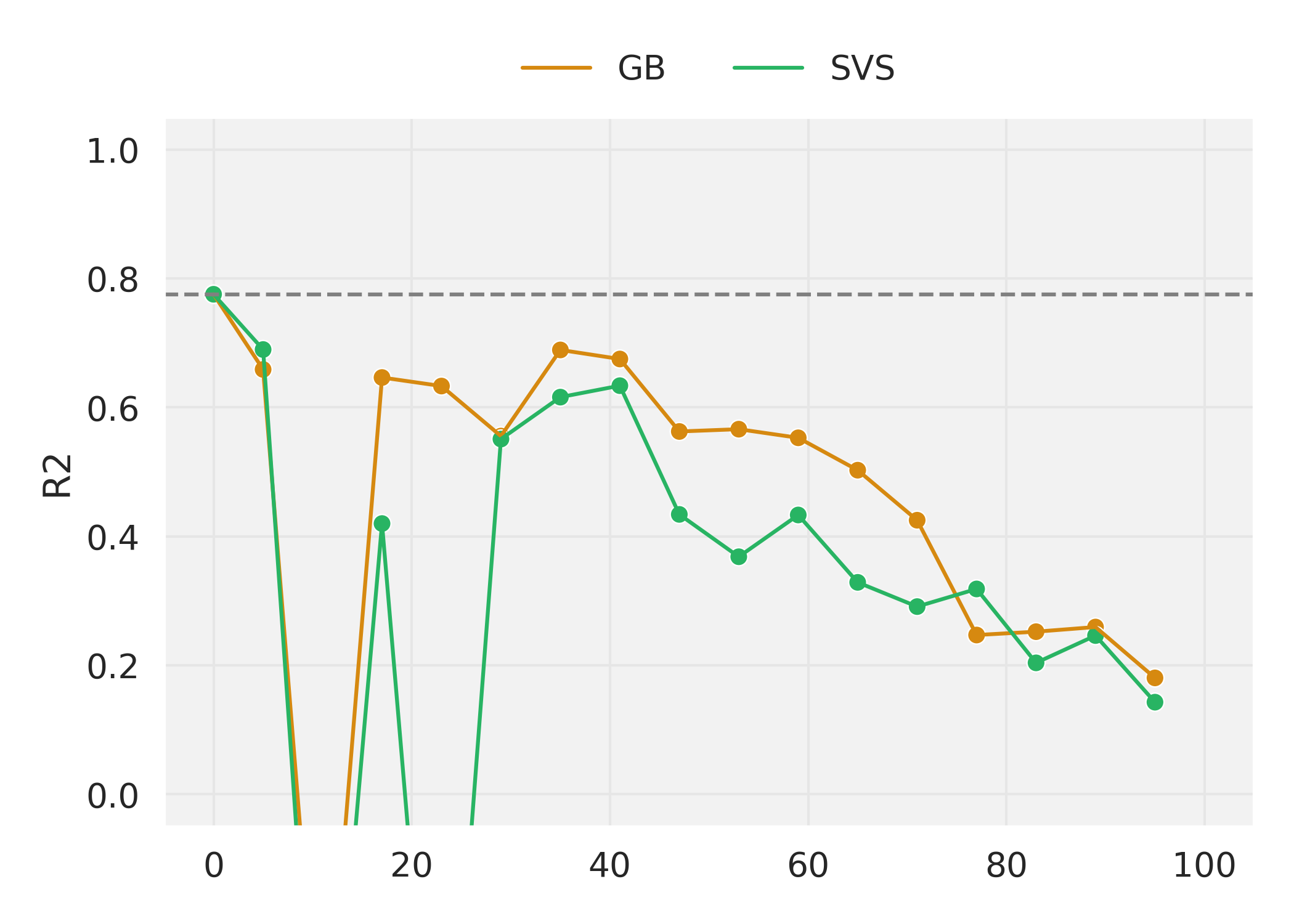}
        \end{subfigure} \\
        \begin{subfigure}{0.3\textwidth}
            \centering
            \includegraphics[width=\linewidth]{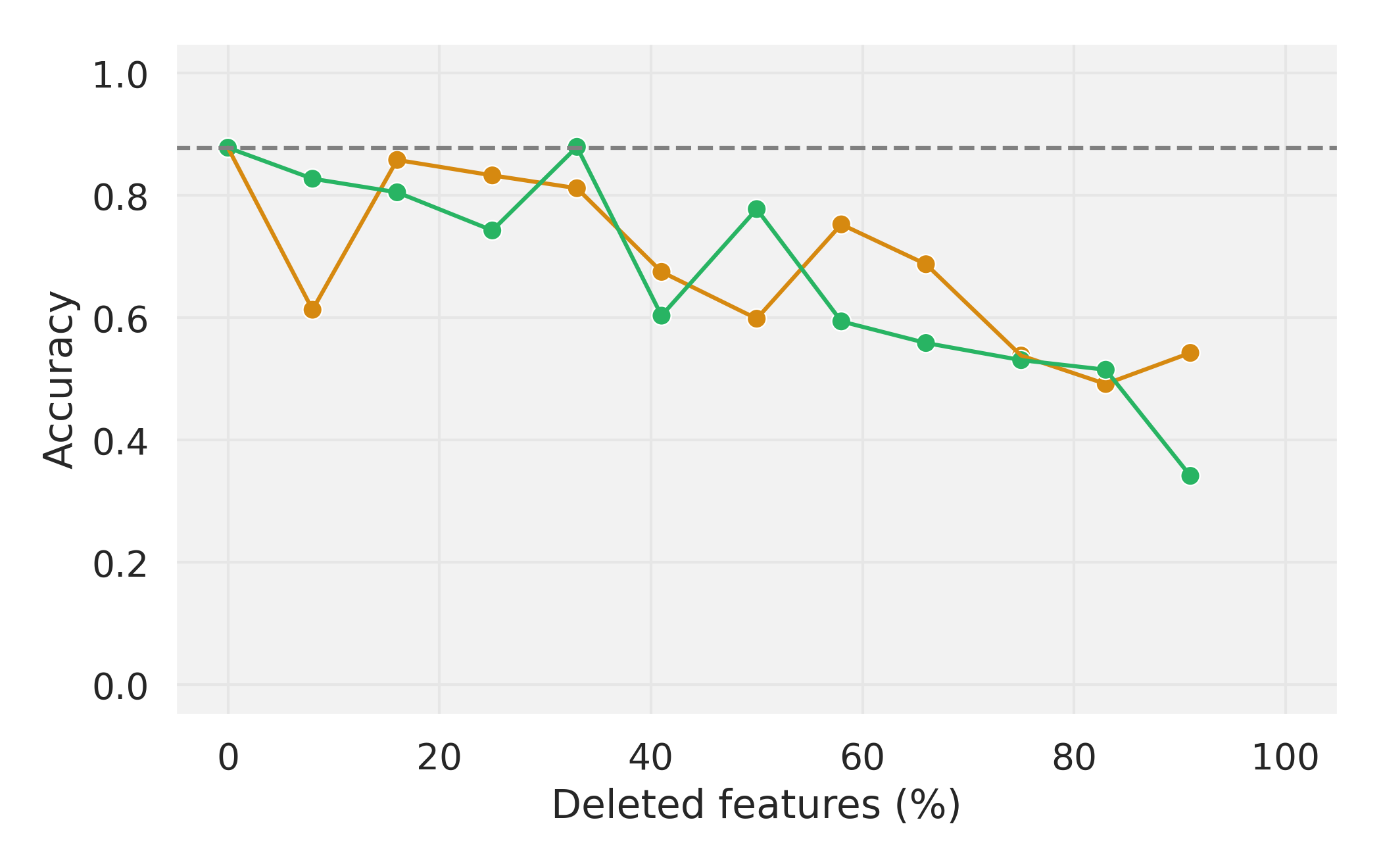}
            \caption{CropHarvest}
        \end{subfigure} &
        \begin{subfigure}{0.3\textwidth}
            \centering
            \includegraphics[width=\linewidth]{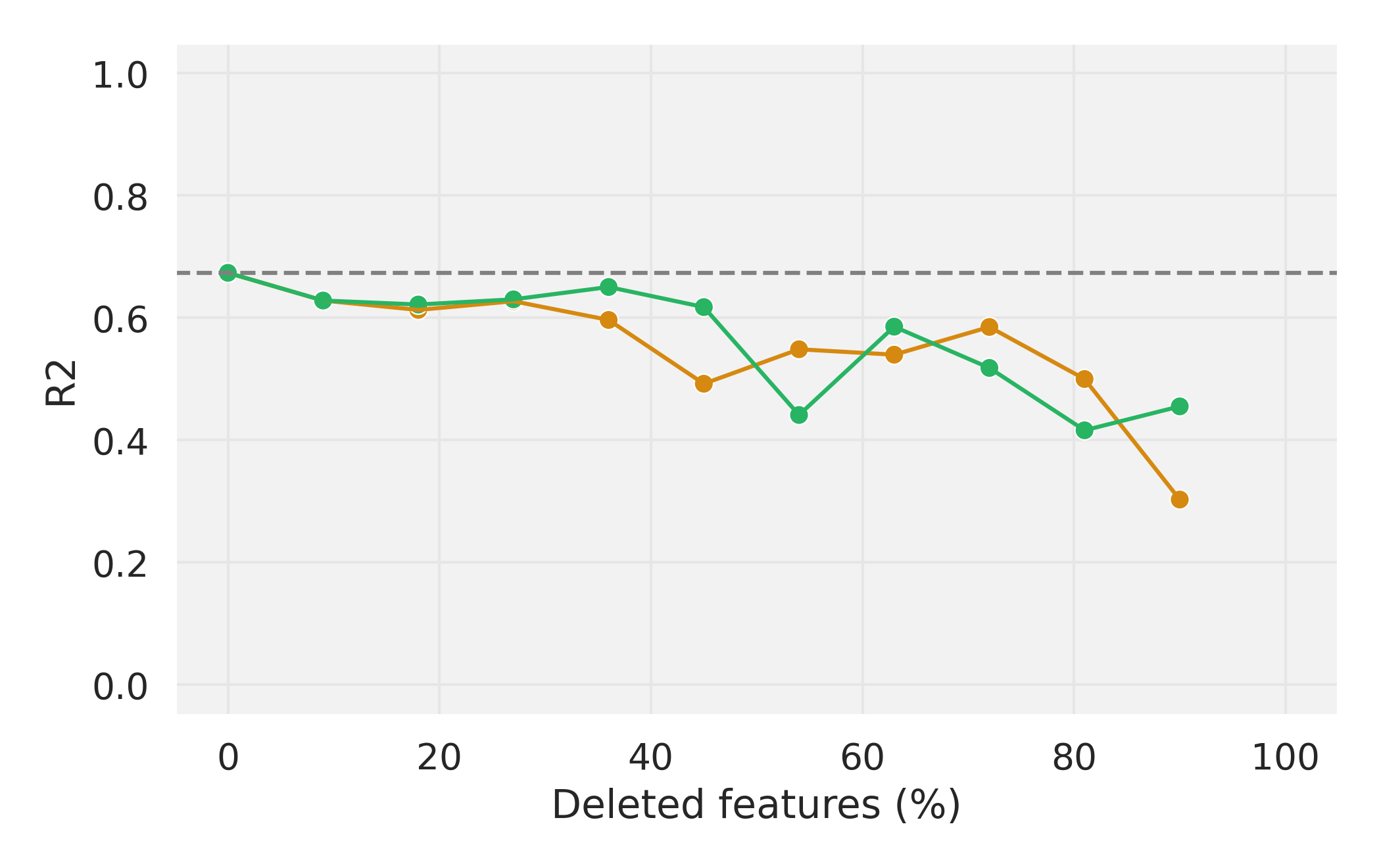}
            \caption{CropYield}
        \end{subfigure} &
        \begin{subfigure}{0.3\textwidth}
            \centering
            \includegraphics[width=\linewidth]{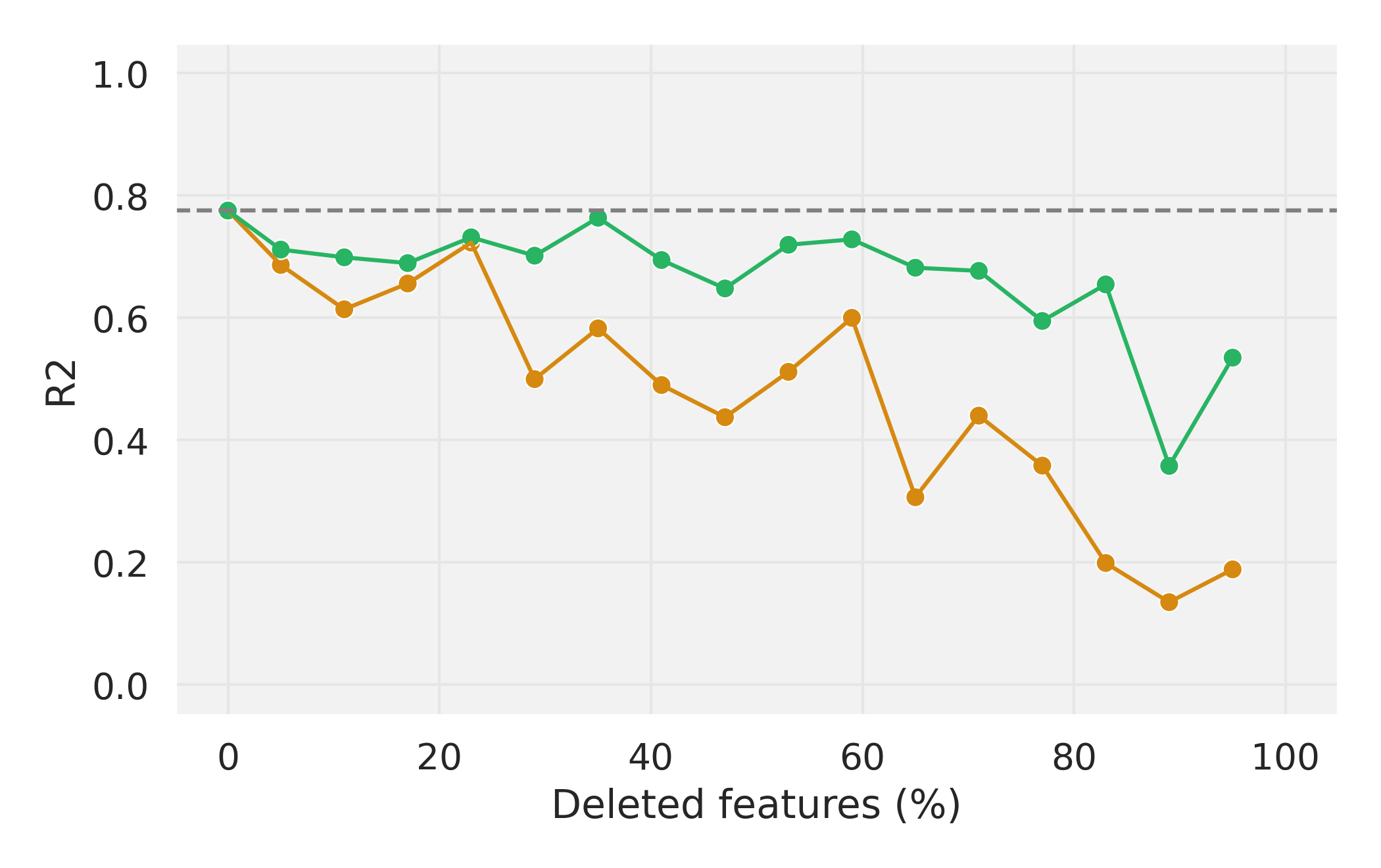}
            \caption{China PM2.5}
        \end{subfigure}
    \end{tabular}
    \caption{Incremental time-step deletion results, comparing  \gls{svs} and \gls{gb} attribution estimators. In the top row the most important time-steps are removed first, and the least important time-steps in the second row.}
    \label{fig:ts_del}
\end{figure}

%% file: tables_figures/band_deletion_figures_sg.tex
\begin{figure}[ht]
    \centering
    \begin{tabular}{ccc}
        \begin{subfigure}{0.3\textwidth}
            \centering
            \includegraphics[width=\linewidth]{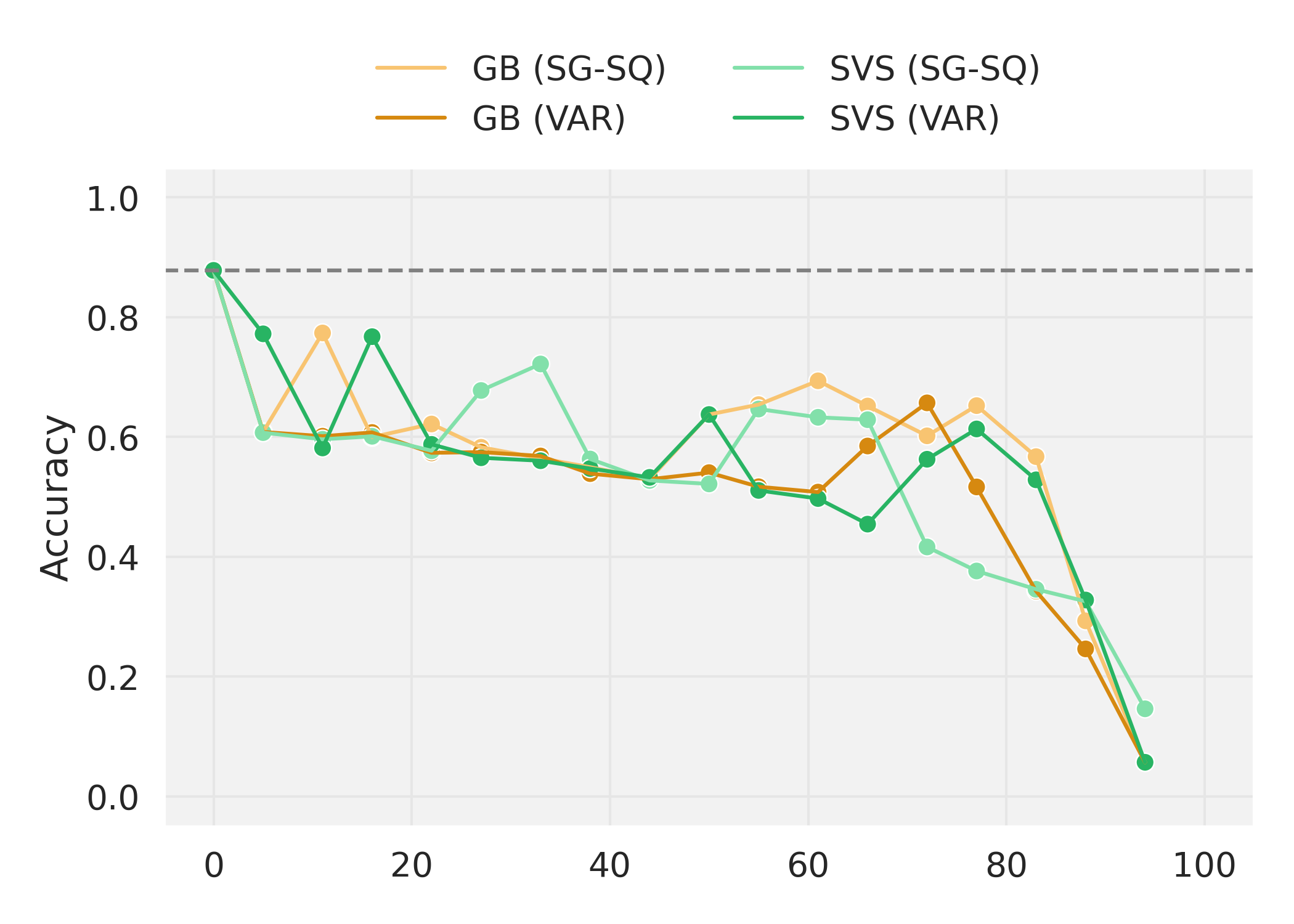}
        \end{subfigure} &
        \begin{subfigure}{0.3\textwidth}
            \centering
            \includegraphics[width=\linewidth]{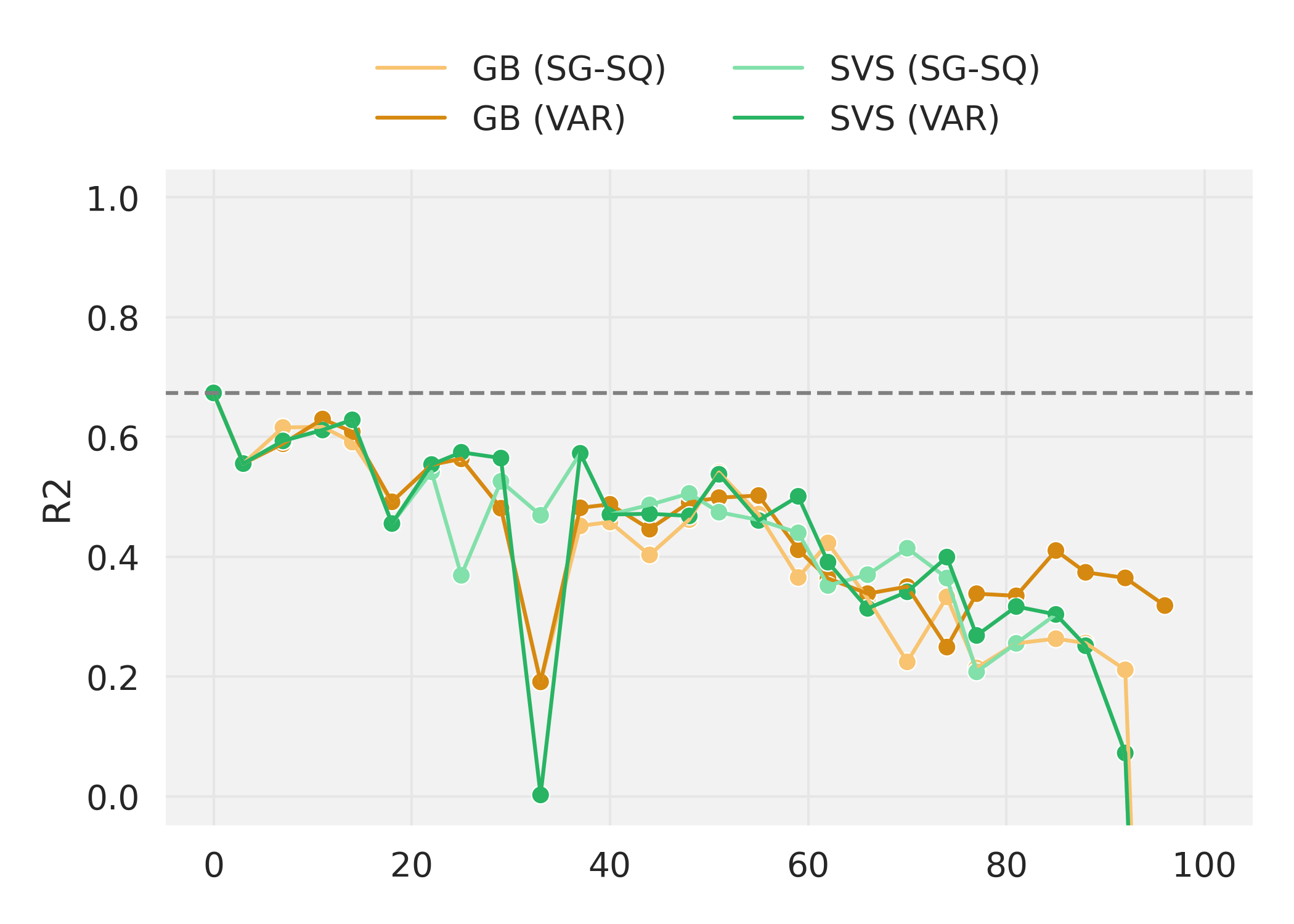}
        \end{subfigure} &
        \begin{subfigure}{0.3\textwidth}
            \centering
            \includegraphics[width=\linewidth]{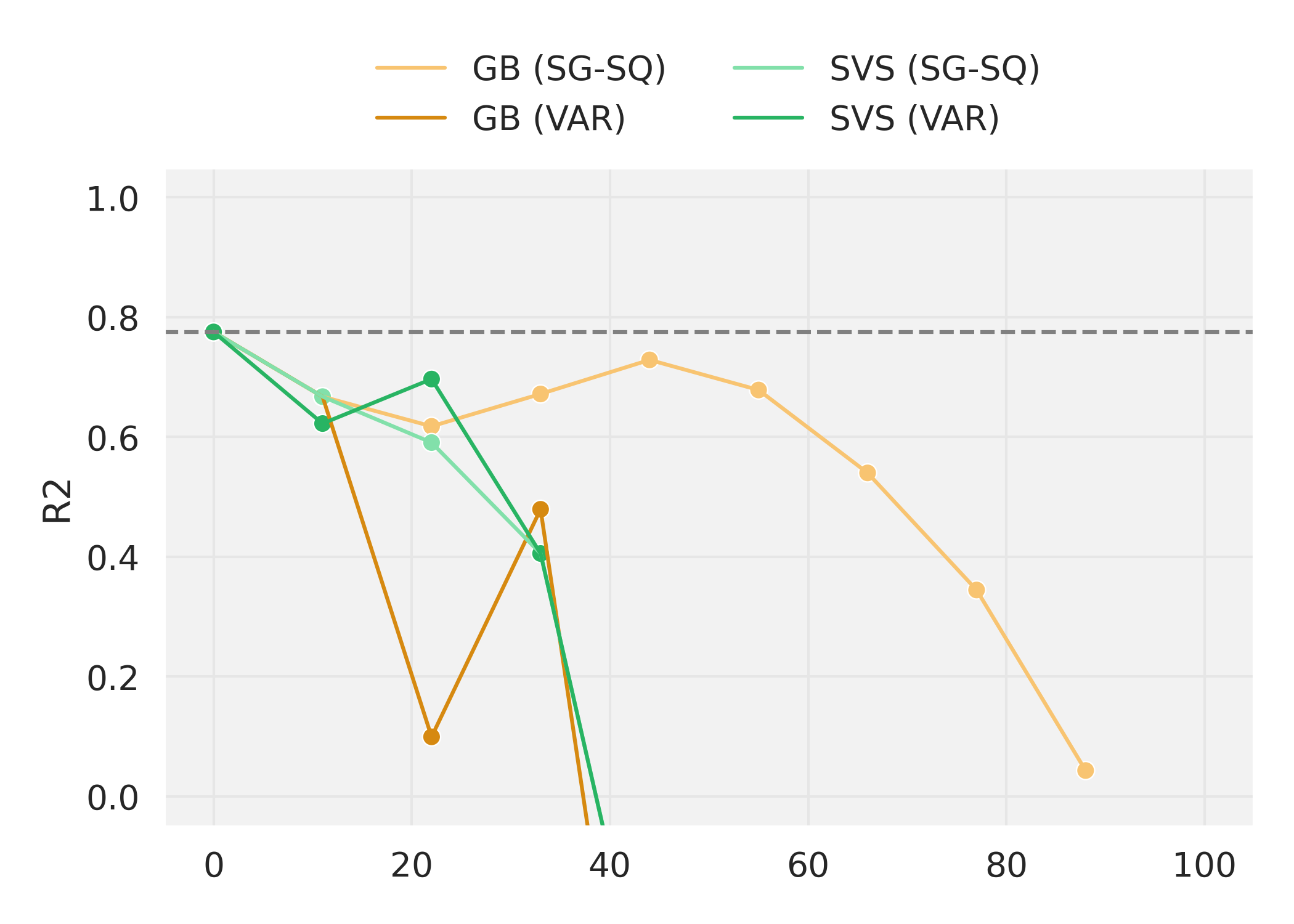}
        \end{subfigure} \\
        \begin{subfigure}{0.3\textwidth}
            \centering
            \includegraphics[width=\linewidth]{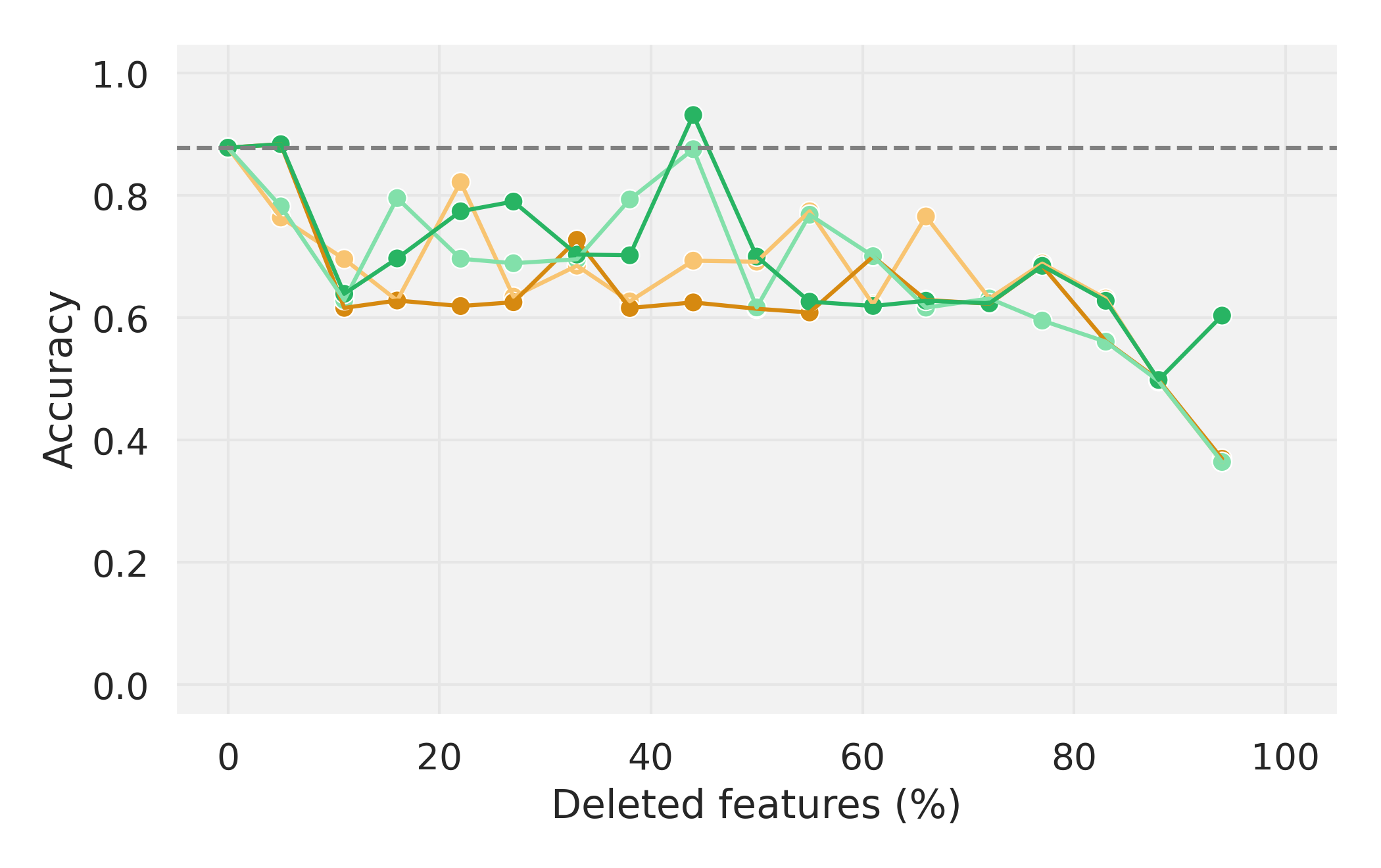}
            \caption{CropHarvest}
        \end{subfigure} &
        \begin{subfigure}{0.3\textwidth}
            \centering
            \includegraphics[width=\linewidth]{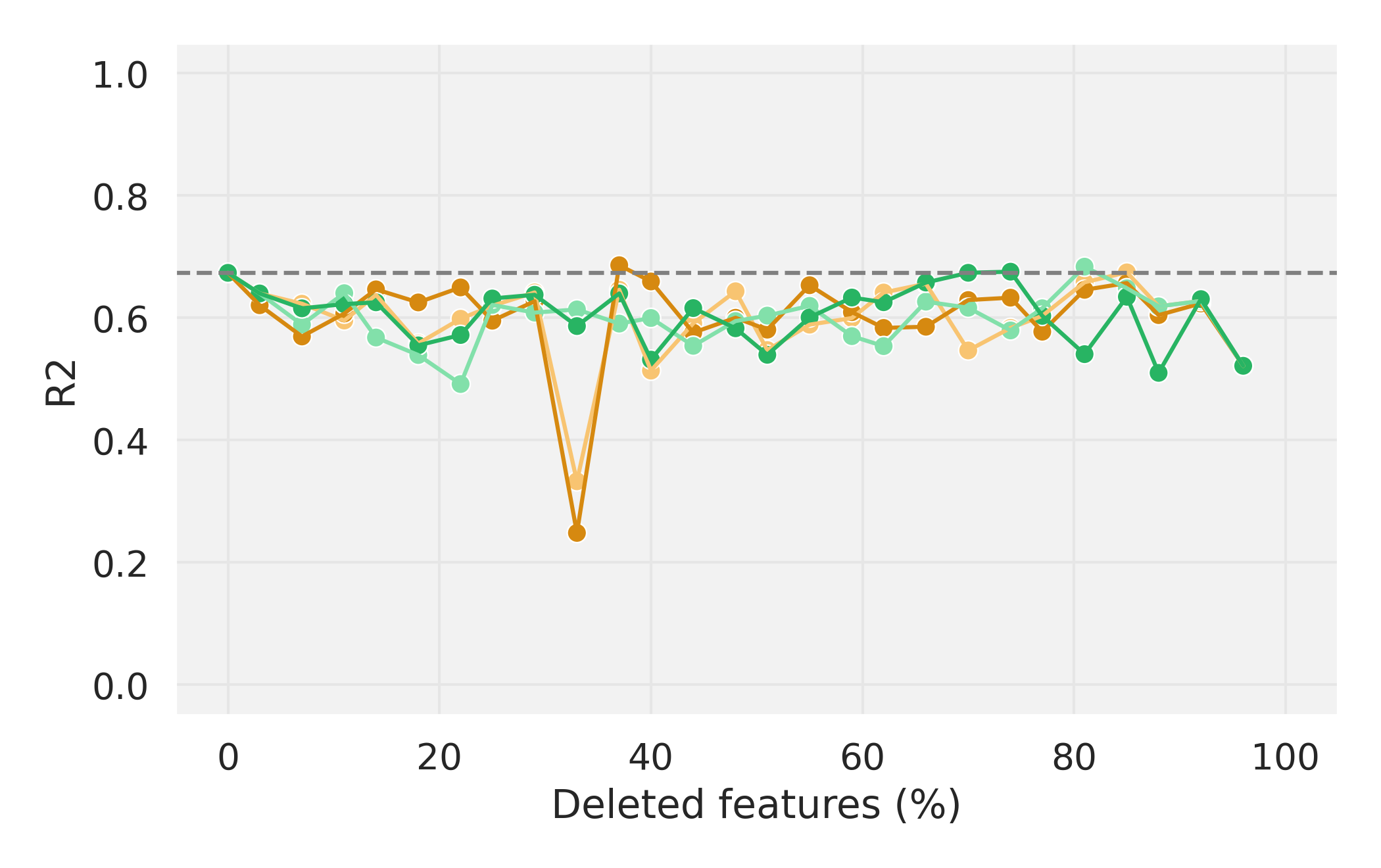}
            \caption{CropYield}
        \end{subfigure} &
        \begin{subfigure}{0.3\textwidth}
            \centering
            \includegraphics[width=\linewidth]{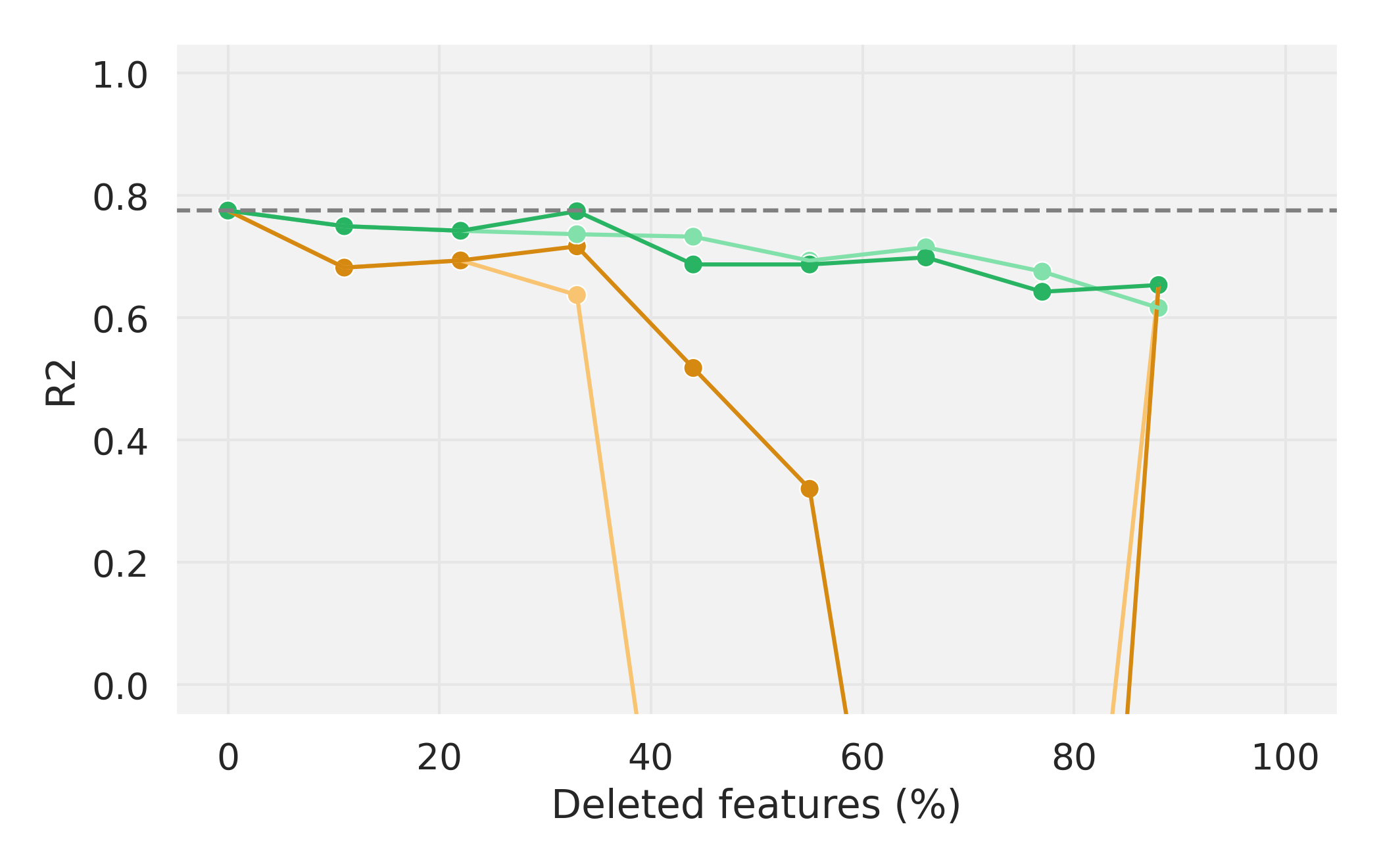}
            \caption{China PM2.5}
        \end{subfigure}
    \end{tabular}
    \caption{Incremental band deletion results,  comparing the ensemble-based estimators \gls{sgs} and \gls{var}, applied to \gls{svs} and \gls{gb}. In the top row the most important bands are removed first, and the least important bands in the second row.}
    \label{fig:band_del_sg}
\end{figure}

%% file: tables_figures/ts_deletion_figures_sg.tex
\begin{figure}[ht]
    \centering
    \begin{tabular}{ccc}
        \begin{subfigure}{0.3\textwidth}
            \centering
            \includegraphics[width=\linewidth]{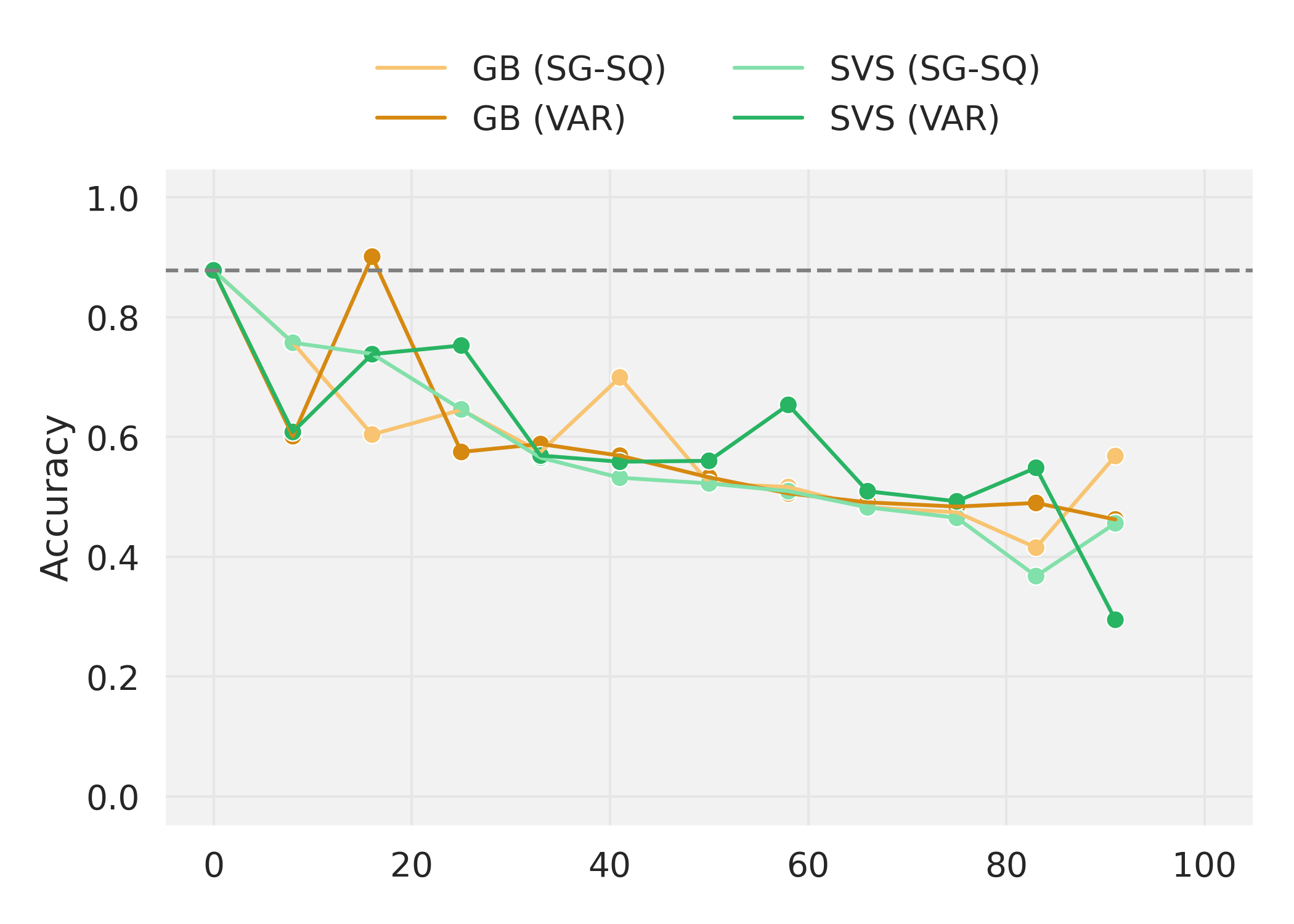}
        \end{subfigure} &
        \begin{subfigure}{0.3\textwidth}
            \centering
            \includegraphics[width=\linewidth]{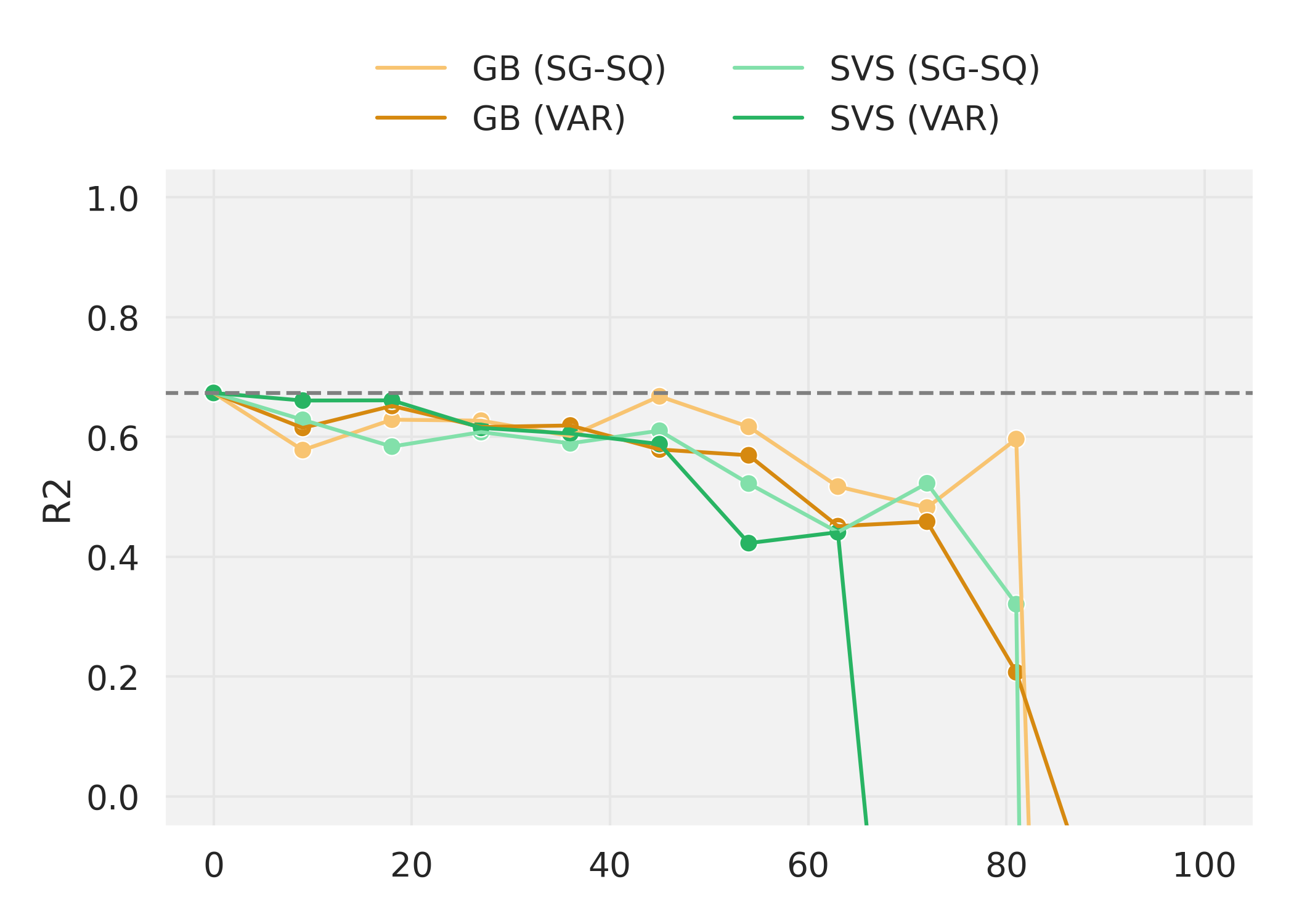}
        \end{subfigure} &
        \begin{subfigure}{0.3\textwidth}
            \centering
            \includegraphics[width=\linewidth]{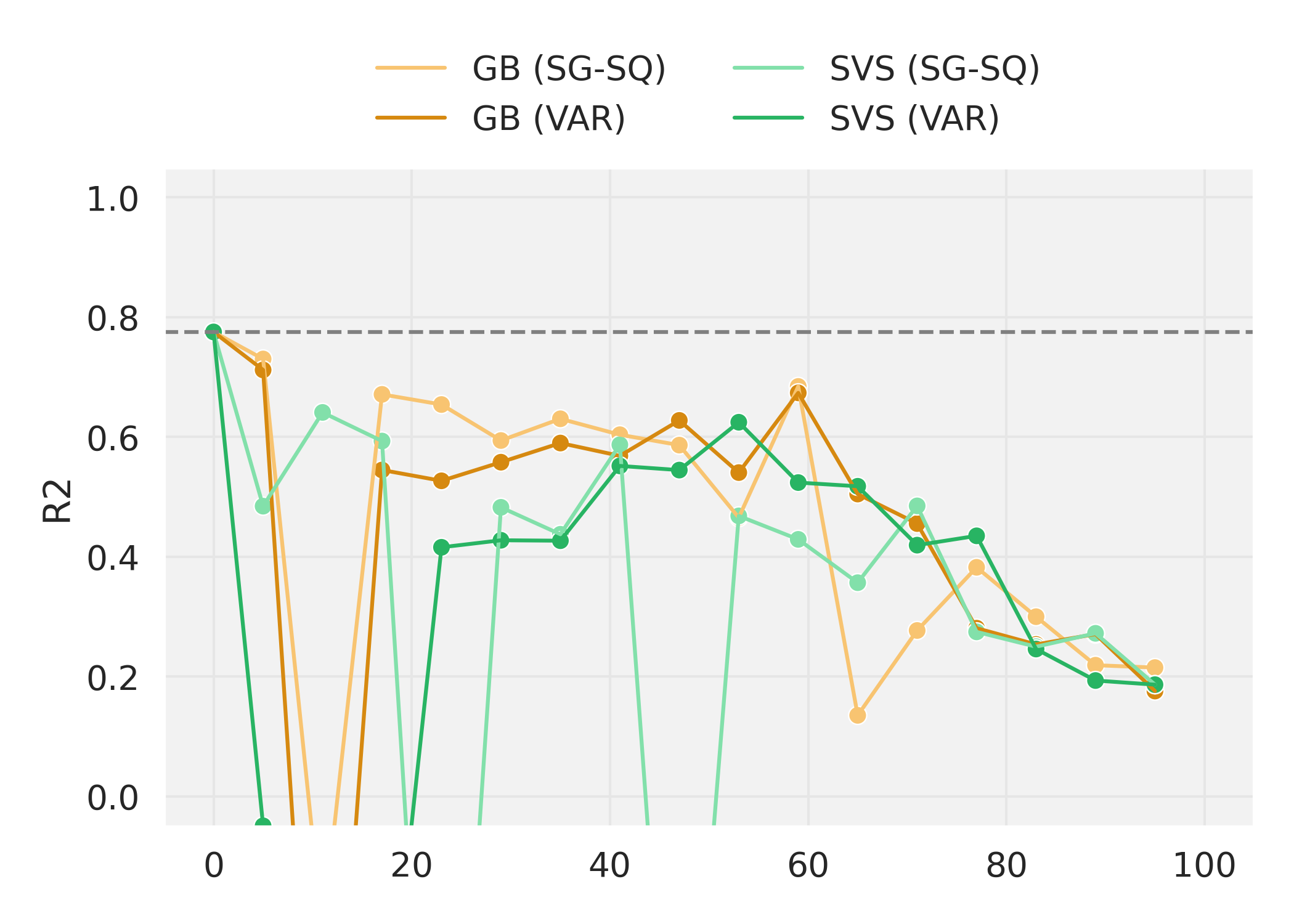}
        \end{subfigure} \\
        \begin{subfigure}{0.3\textwidth}
            \centering
            \includegraphics[width=\linewidth]{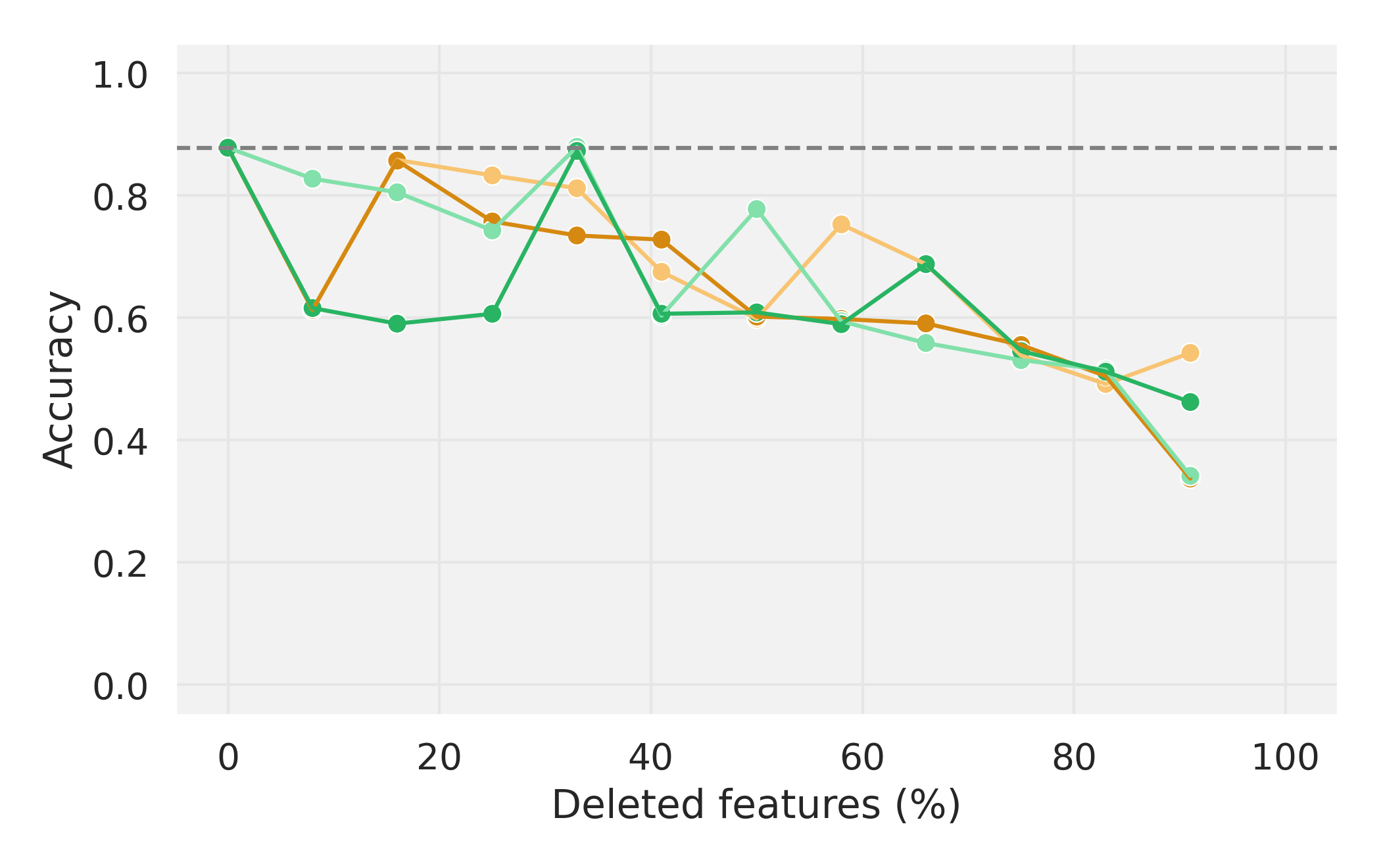}
            \caption{CropHarvest}
        \end{subfigure} &
        \begin{subfigure}{0.3\textwidth}
            \centering
            \includegraphics[width=\linewidth]{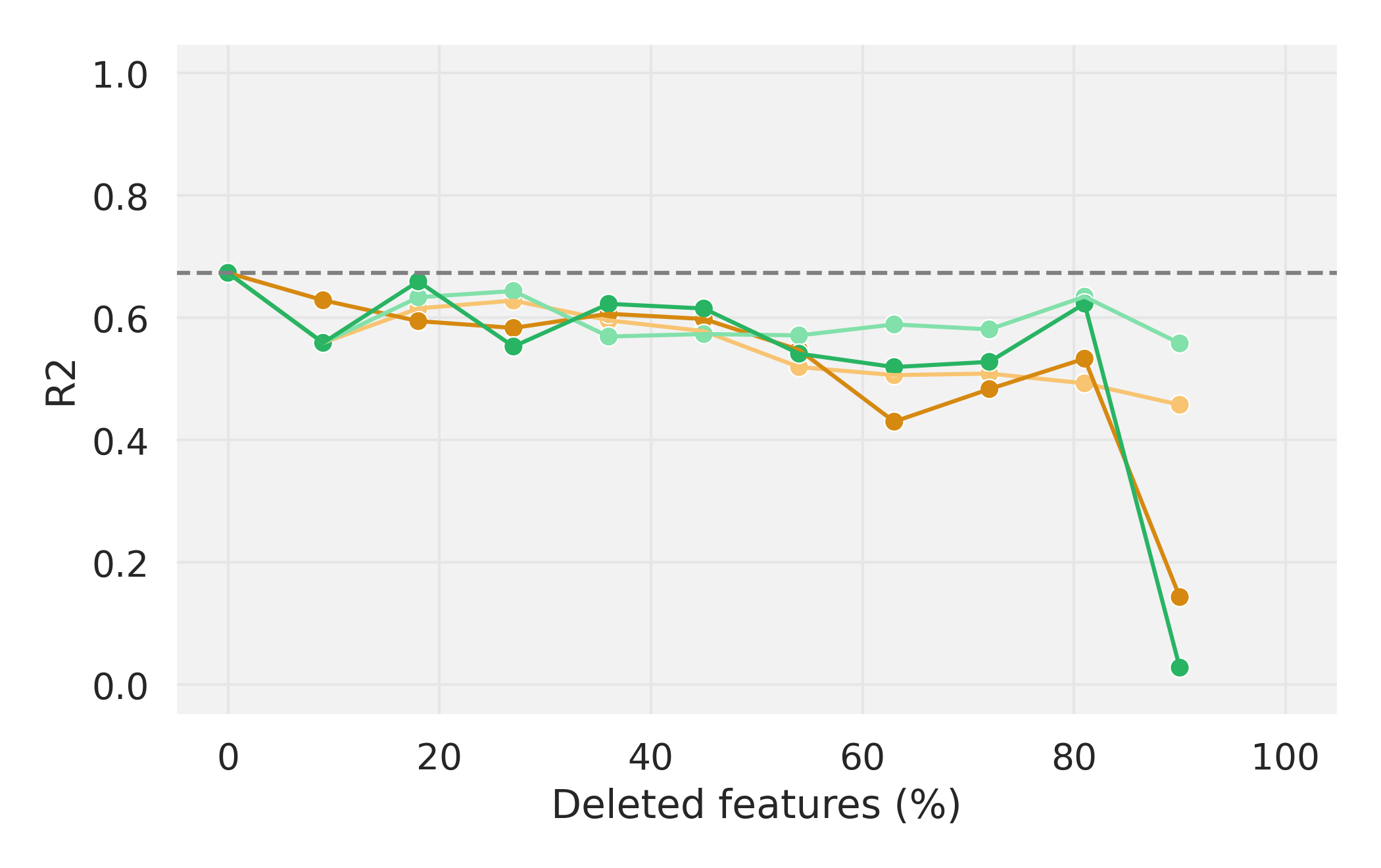}
            \caption{CropYield}
        \end{subfigure} &
        \begin{subfigure}{0.3\textwidth}
            \centering
            \includegraphics[width=\linewidth]{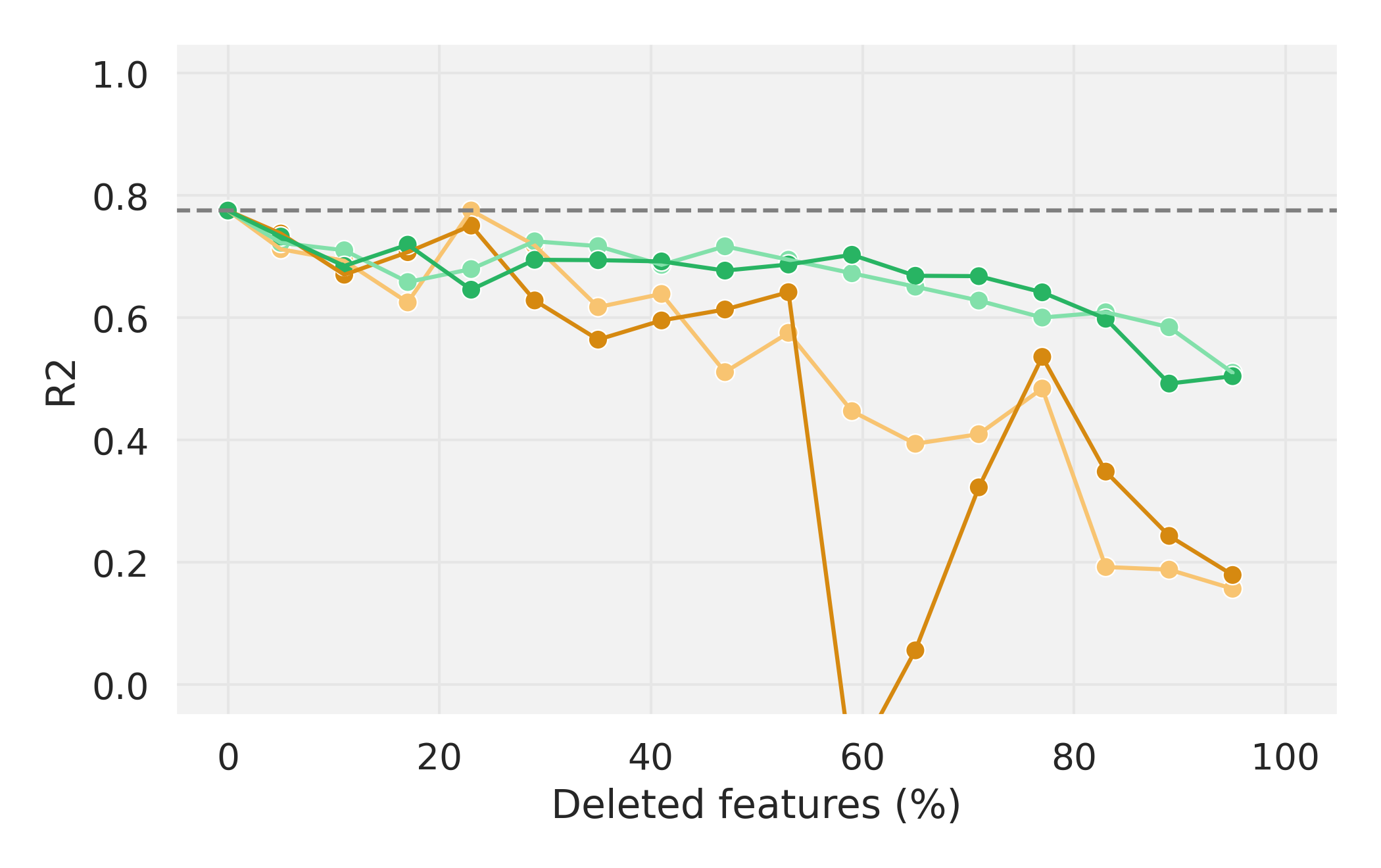}
            \caption{China PM2.5}
        \end{subfigure}
    \end{tabular}
    \caption{Incremental time-step deletion results,  comparing the ensemble-based estimators \gls{sgs} and \gls{var}, applied to \gls{svs} and \gls{gb}. In the top row the most important time-steps are removed first, and the least important time-steps in the second row.}
    \label{fig:ts_del_sg}
\end{figure}

%% file: sections/04_conclusion.tex
\section{Conclusion}\label{sec:conc}

    Inspired by the \gls{roar} framework, we propose an approach to identify a small subset of bands and time-steps in geospatial temporal data sufficient to reach the model's baseline performance, i.e. the accuracy reached when providing the model with all available modalities and instances. 
    We evaluated this approach on three datasets, and showed how many features can be removed before a significant drop in accuracy is observed.
    Additionally, we found that in some datasets, performance declines immediately after a few features identified as the most important are removed. This suggests that these features are necessary for the baseline performance and that the information they encode is absent in the remaining features.
    
    Furthermore, the expected behavior regarding the decline in performance when starting with the deletion of the most or least important features also revealed a higher correctness of the attributions estimated by \gls{svs} compared to \gls{gb}. The ensemble-based variants only improved the faithfulness of \gls{gb} estimates in a few cases.

    This work can be enhanced by comparing additional feature attribution methods. The faithfulness of the chosen method enhances its ability to identify a minimal feature subset necessary to achieve the model's baseline performance. Moreover, comparing these results across various model architectures might help identify the features necessary and sufficient for predicting the target regardless of the model employed.

%% file: sections/99_credits.tex
\begin{credits}
    \subsubsection{\ackname} 
        H.N. acknowledges support through a scholarship from the University of Kaiserslautern-Landau.
    
    \subsubsection{\discintname}
        The authors have no competing interests to declare that are
    relevant to the content of this article.
\end{credits}